\documentclass[sigconf]{acmart}

\AtBeginDocument{%
  \providecommand\BibTeX{{%
    \normalfont B\kern-0.5em{\scshape i\kern-0.25em b}\kern-0.8em\TeX}}}

\setcopyright{rightsretained}
\copyrightyear{2020}

\acmConference[]{}{}

\usepackage[ruled,vlined,linesnumbered]{algorithm2e}
\usepackage{bm}
\usepackage{colortbl} 
\usepackage{algorithmic}
\usepackage{multirow}
\usepackage{bbding}
\usepackage{subcaption}
\usepackage{amsthm}

\newtheorem{dfn}{Definition}[section]

\newcommand{\cm}{\Checkmark}
\newcommand{\xm}{\XSolidBrush}
\newcommand{\argmax}{\mathop{\rm arg~max}\limits}

\begin{document}


\title{One-vs.-One Mitigation of Intersectional Bias: A General Method to Extend Fairness-Aware Binary Classification}

\author{Kenji Kobayashi}
\email{kobayashi_kenji@fujitsu.com}
\affiliation{%
  \institution{Fujitsu Laboratories Ltd.}
  \city{Kawasaki}
  \country{Japan}
}
\author{Yuri Nakao}
\email{nakao.yuri@fujitsu.com}
\affiliation{%
  \institution{Fujitsu Laboratories Ltd.}
  \city{Kawasaki}
  \country{Japan}
}

\renewcommand{\shortauthors}{Kobayashi and Nakao}
\renewcommand{\shorttitle}{One-vs.-One Mitigation of Intersectional Bias}
\begin{abstract}
With the widespread adoption of machine learning in the real world, the impact of the discriminatory bias has attracted attention. In recent years, various methods to mitigate the bias have been proposed. However, most of them have not considered intersectional bias, which brings unfair situations where people belonging to specific subgroups of a protected group are treated worse when multiple sensitive attributes are taken into consideration. 
To mitigate this bias, in this paper, we propose a method called \textbf{One-vs.-One Mitigation} by applying a process of comparison between each pair of subgroups related to sensitive attributes to the fairness-aware machine learning for binary classification. 
We compare our method and the conventional fairness-aware binary classification methods in comprehensive settings using three approaches (pre-processing, in-processing, and post-processing), six metrics (the ratio and difference of demographic parity, equalized odds, and equal opportunity), and two real-world datasets (Adult and COMPAS). As a result, our method mitigates the intersectional bias much better than conventional methods in all the settings. With the result, we open up the potential of fairness-aware binary classification for solving more realistic problems occurring when there are multiple sensitive attributes.

\end{abstract}





\maketitle

\begin{figure}[t]
  \begin{center}
    \includegraphics[width=\linewidth]{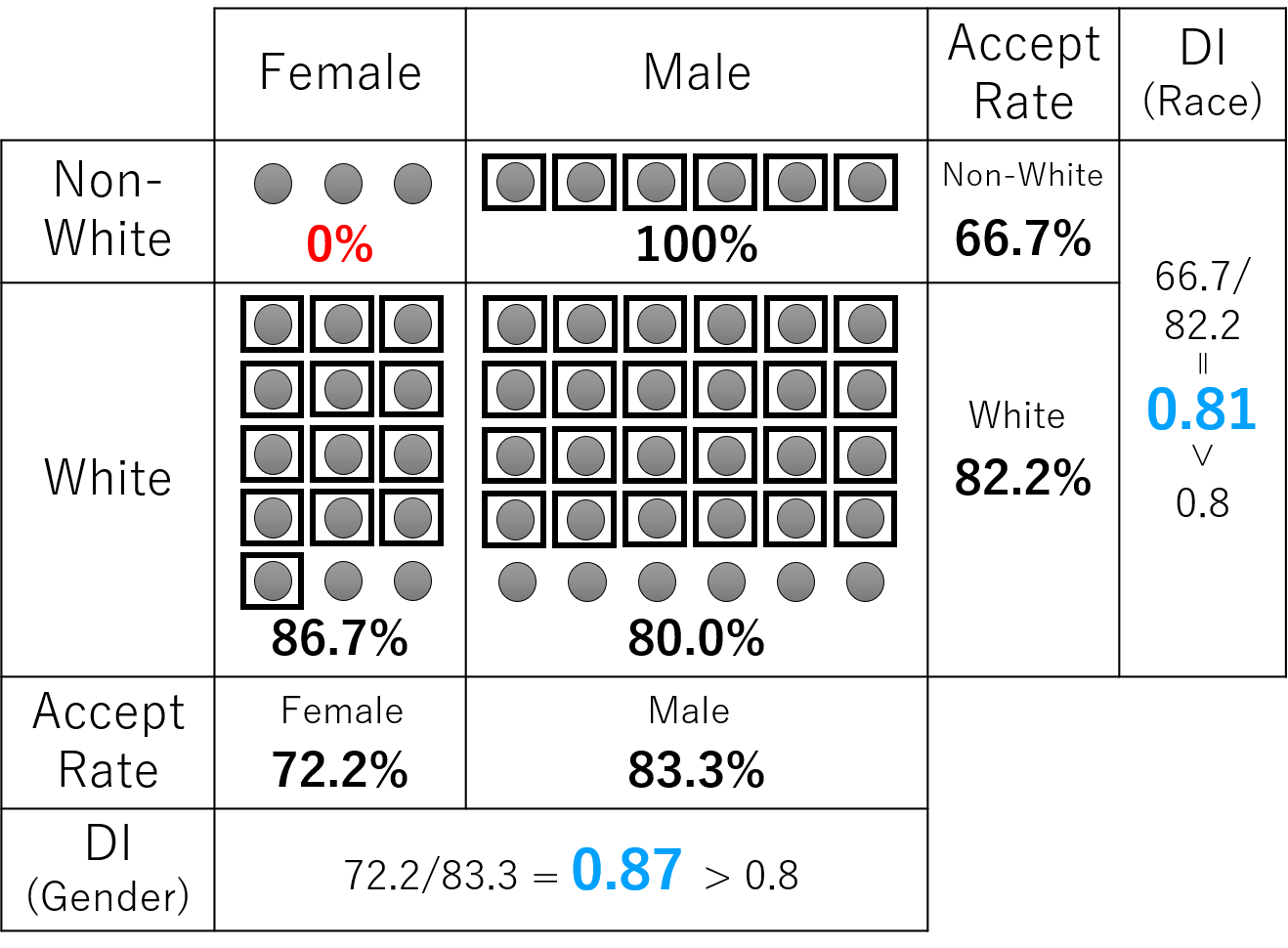}
    \caption{A toy example of intersectional bias. Grey circles are the applicants for loan applications, and those surrounded by black squares are the accepted ones. The percentages are the acceptance rate in each group. In this example, the disparate impact (DI), which is the ratio of the acceptance rate of a protected group to that of a non-protected group is used as the fairness metrics. In US law, it is said that if the value of disparate impact is more than 0.8, there is not an unfair situation (80\% rule)~\cite{10.1145/2783258.2783311}. However, in this example, Even if the fairness metrics are satisfied in each sensitive attribute (DI is more than 0.8 for both gender and race), there is a subgroup that is clearly discriminated, non-white female, whose acceptance rate is 0\%.}
    \label{fig:IntesectionalBias}
  \end{center}
\end{figure}

\section{Introduction}
\label{sec:intro}

The issue of fairness in machine learning technology is currently gaining more attention, and various technologies \cite{10.1007/s10115-011-0463-8, 10.1145/2783258.2783311, NIPS2017_6988, 10.1145/3278721.3278779, 10.1007/978-3-642-33486-3_3, 10.1145/3287560.3287586, 6413831, NIPS2016_6374} have been developed to solve this. 
This issue occurs because machine learning models used in diverse decision makings, such as loan applications or bail decisions are often trained with dataset including discriminatory bias that human decision-makers have had, e.g., bias based on gender or race.
To overcome the issue, fairness-aware machine learning to mitigate the bias in the training data~\cite{10.1007/s10115-011-0463-8, 10.1145/2783258.2783311, NIPS2017_6988} or machine learning models~\cite{10.1145/3278721.3278779, 10.1007/978-3-642-33486-3_3, 10.1145/3287560.3287586, 6413831, NIPS2016_6374} have been developed. 
Especially, for binary classification tasks, these technologies have been explored focusing on handling diverse fairness metrics, such as demographic parity~\cite{Barocas-BigDataDI-2016}, equalized odds, or equal opportunity~\cite{NIPS2016_6374, Zafar-2017-FairnessBeyondDTDI} in accordance with the purpose.

Despite various fairness criteria are considered with various methods as described above, most of them have not considered intersectional bias~\cite{gendershadespmlr-v81-buolamwini18a, Cabrera-fairvis-2019} that occurs in the practical situation where there are multiple sensitive attributes. 
This bias is such that even when a protected group seems to be treated fairly as a whole, a subset of the protected group can be treated unfairly.
As an example, consider the case where there are two binary sensitive attributes: races (non-white and white), and genders (female and male) (Figure~\ref{fig:IntesectionalBias}).
When we try to provide similar treatment for both races and genders respectively, although the chosen fairness criterion (disparate impact~\cite{10.1145/2783258.2783311}) is satisfied on all sensitive attributes independently, there is a subgroup that is discriminated clearly, such as the non-white female group in the case in Figure~\ref{fig:IntesectionalBias}.
Because the most conventional methods attempted to mitigate the bias based on only one sensitive attribute, they have not considered the issue of the intersectional bias.

Additionally, intersectional bias should be mitigated in any use-case scenario because the bias can exist in any situation of decision making.
In fact, there are several conventional methods to deal with the intersectional bias or similar issues called subgroup fairness or fairness gerrymandering~\cite{pmlr-v80-hebert-johnson18a, pmlr-v80-kearns18a, 10.1145/3287560.3287592, foulds2020intersectional, Multiaccuracy10.1145/3306618.3314287}. 
However, it is still difficult to apply them to diverse use-case scenarios because they focused on introducing their own metrics or limited use-case scenarios and mitigating bias on the basis of them.

In this paper, we propose a general method to enable any fairness-aware binary classification method to mitigate intersectional bias in diverse use-case scenarios. 
By applying a process of comparison between each pair of subgroups related to sensitive attributes to the conventional fairness-aware methods of binary classification, our method extends the conventional methods regardless of the approach they take.
Our method calculates a score for each instance (i.e., a data of an applicant) and searches for an appropriate threshold of the score that divides a dataset into favorable and unfavorable classes. In this paper, we treat a favorable class as a positive class, and an unfavorable class as a negative class.
The threshold is determined differently for the different subgroups considering the trade-off between accuracy and to what extent the bias is mitigated, which can be decided by users.
We experimented with three classic fairness criteria and four conventional methods that cover three approaches of fairness-aware machine learning on two real-world datasets.

Our contributions are as follows:
\begin{enumerate}
\item Our method enables intersectional bias mitigation while inheriting the fairness criteria and approach types supported by conventional methods to meet the wide range of requirements from analysts and decision-makers.
\item Our method provides a subgroup disparity upper limit, which can control the trade-off between accuracy and fairness.
\item We demonstrate that our method can cover diverse use-case scenarios of decision makings using two real-world datasets.
\end{enumerate}

The remaining of this paper is organized as follows:
we begin to discuss related works about fairness-aware machine learning and intersectional bias in Section \ref{section:related_work}.
In Section \ref{section:preliminaries}, we provide preliminaries such as notation, fairness criteria.
In Section \ref{section:method}, we propose a general method called one-vs.-one mitigation, which enables bias mitigation technologies to deal with intersectional bias.
In Section \ref{sec:experiment}, we set up an experimental method to measure the effect of our proposed method on the intersectional bias.
In Section \ref{section:results}, we demonstrate the experimental result with three approaches, six metrics, and two real-world datasets.
In Section \ref{section:discussion}, we discuss the effect of intersectional bias mitigation on disparity and accuracy based on the result.
Finally, in Section \ref{section:conclusion}, we describe the conclusion of this paper.

\section{Related work}\label{section:related_work}

In this section, we describe the related works of fairness-aware machine learning and focus on those related to intersectional bias. 

\subsection{Fairness-Aware Binary Classification}
In the field of binary-classification of fairness-aware machine learning, various methods have been proposed to mitigate discriminatory bias from the results in the situation where there is a single sensitive attribute.
The methods are categorized into three approaches: pre-, in-, and post-processing.
Pre-processing is used to mitigate the discriminatory bias in the training data~\cite{10.1145/2783258.2783311, 10.1007/s10115-011-0463-8} or both training and test data~\cite{NIPS2017_6988} by, for example, modifying the class label~\cite{10.1007/s10115-011-0463-8}, feature values~\cite{10.1145/2783258.2783311}.
In-processing approach methods mitigate the bias while training the models by introducing fairness constraint terms based on specific fairness criteria~\cite{10.1145/3278721.3278779, 10.1007/978-3-642-33486-3_3, 10.1145/3287560.3287586}.
Post-processing approach methods mitigate the bias in the results from basic machine learning models~\cite{6413831, NIPS2016_6374} e.g., by introducing the mitigation model after a plain classifier~\cite{NIPS2016_6374, 6413831}. 
With the above methods, diverse bias mitigation tasks have been executed.
However, most conventional methods have not considered intersectional bias, which occurs when there are multiple sensitive attributes. 

Additionally, Each approach has its advantages and disadvantages.
The pre-processing approach\cite{10.1145/2783258.2783311, 10.1007/s10115-011-0463-8, NIPS2017_6988} is advantageous in terms of privacy because users do not need to use the sensitive attributes when using classifiers. On the other hand, because this approach mitigates the bias only in datasets, not in models, it cannot handle the metrics related to accuracy.
For the in-processing approach~\cite{10.1145/3278721.3278779, 10.1007/978-3-642-33486-3_3, 10.1145/3287560.3287586}, it is an advantage that the approach prevent the trade-off between fairness and accuracy, which exists in other approaches. However, with this approach, every time the machine learning task changes, the classifier itself needs to be modified, which takes a lot of time and effort.
Finally, post-processing methods\cite{6413831, NIPS2016_6374} are advantageous in that they have simple structures because they do not use non-sensitive attributes when mitigating the bias. 
However, they do not tend to achieve as high performance as in-processing approach methods because what they can handle is restricted to the sensitive attributes when mitigating the bias.

As described above, there are not any one-size-fits-all approach, and users have to choose appropriate one considering these advantages and disadvantages.
Considering this situation, in this paper, we propose a method to mitigate intersectional bias with all of these approaches to enable the users to consider the intersectional bias in any use-case scenario.

\subsection{Fairness Metrics}
To evaluate unfair situations, various group-based~\cite{Barocas-BigDataDI-2016, NIPS2016_6374, Zafar-2017-FairnessBeyondDTDI, Russel-2017-WorldsCollide} and individual-based~\cite{Dwork-2012-Awareness} fairness metrics have been considered.
Here, we focus on the group-based metrics because intersectional bias occurs when we consider the discrimination of protected groups related to sensitive attributes.
As the group-based fairness metrics, demographic parity~\cite{Barocas-BigDataDI-2016}, equalized odds, equal opportunity \cite{NIPS2016_6374, Zafar-2017-FairnessBeyondDTDI}, or counterfactual fairness\cite{Russel-2017-WorldsCollide} are considered in existing studies.
Among them, most conventional studies have used metrics related to demographic parity or ones to error rate (i.e., equalized odds or equal opportunity).
This is because these metrics can be flexibly applied to diverse situations.
For example, to ignore the relationship between sensitive attributes and the outputs for the purpose of affirmative action, demographic parity is appropriate. 
On the other hand, if analysts try to consider the decisions in actual data, they will use the metrics related to the error rate, equalized odds, or equal opportunity.

In this paper, we pick up the demographic parity, equalized odds, and equal opportunity as the most applicable fairness metrics with which our method is evaluated.
This is because we aim at enabling the users to take the intersectional bias into account in diverse contexts with our method.

\subsection{Intersectional Bias}

Intersectional bias has its roots in the sociological concept, `\emph{Intersectionality}'~\cite{crenshaw1989demarginalizing}.
The concept of \emph{Intersectionality} covers diverse discussions including the issue of the oppression that people feel due to the discrimination~\cite{foulds2020intersectional}.
In a monumental paper published in 1989, Kimberl\'{e} Crenshaw~\cite{crenshaw1989demarginalizing} introduced \emph{Intersectionality} by referencing a court case where black women were unfairly discriminated as a result of an activity to mitigate the race and gender discrimination independently.

By treating all people in any subset of protected groups fairly, i.e., by guaranteeing fairness in terms of intersectional bias, machine learning technology can contribute to addressing the issue of \emph{Intersectionality}. 
So far, some conventional studies have addressed the issue of intersectional bias~\cite{gendershadespmlr-v81-buolamwini18a, Cabrera-fairvis-2019}. 
Buolamwini and Gebru~\cite{gendershadespmlr-v81-buolamwini18a} shed light on intersectional bias in commercial gender classification systems with a new facial image dataset balanced by gender and skin type.
Cabrera et al.~\cite{Cabrera-fairvis-2019} developed a visual analytics system for discovering intersectional bias called FairVis. 

On the other hand, several conventional studies of fairness-aware machine learning have considered similar issues to intersectional bias~\cite{pmlr-v80-hebert-johnson18a, pmlr-v80-kearns18a, 10.1145/3287560.3287592, foulds2020intersectional, Multiaccuracy10.1145/3306618.3314287}.
Kearns et al.~\cite{pmlr-v80-kearns18a} proposed a new concept of `subgroup fairness.' 
Their concept is similar to but different from the fairness in intersectional bias because, in their concept, the smaller the size of the concerned subgroup, the smaller the evaluated extent to which the subgroup is unfairly treated, which leads to ignoring the discrimination for the minority group.
To overcome the issue of ignorance of the minority group and consider intersectionality, Foulds et al.~\cite{foulds2020intersectional} proposed a new metric based on statistical parity. 
However, because their metric is designed to work with their in-process algorithm, the range of its application to diverse situations is restricted. 
Additionally, there have been studies that attempted to guarantee accurate results in situations where there are diverse subgroups with regression~\cite{pmlr-v80-hebert-johnson18a} and binary classification tasks~\cite{Multiaccuracy10.1145/3306618.3314287}.
Although they achieved results without a fairness-utility trade-off, their approaches are not applicable in the context where demographic parity, which ignores the accuracy, should be used.
Therefore, to the best of our knowledge, there are no applicable approaches to mitigate intersectional bias to the diverse contexts that fairness-aware machine learning can deal with in general. 

In this paper, we propose a general method to mitigate intersectional bias that is applicable to as diverse approaches and criteria as possible.
For this purpose, rather than introducing new criteria to measure intersectional bias, we develop technology to mitigate intersectional bias based on conventional criteria.

\section{preliminaries}\label{section:preliminaries}

We consider binary classification tasks.
We take a dataset whose instance $Xi$ includes a true class {$C$}, a determined class {$Z$}, sensitive attributes {$S$}, and non-sensitive attributes {$A$}. 
$Z$ is the class obtained as a result of a specific processing (e.g., prediction, mitigation).
{$ C $} and {$ Z $} are binary classes $\{+,-\}$, where {$+$} is a favorable class (e.g., accepted in loan applications, passing recruitment examinations), whereas {$-$} is an unfavorable class.
Hereinafter, we consider a favorable class as a positive class and an unfavorable class as a negative class. 
We assume that the sensitive attributes {$ S \in \{\mathcal{S}^1 \times \ldots \times \mathcal{S}^l\} $}, and non-sensitive attributes {$ A \in \{\mathcal{A}^1 \times \ldots \times \mathcal{A}^m\} $} have multiple attributes (e.g., race, gender, age) and polyvalent attributes (e.g., race, those that have more than two values such as Blue, Green, and Purple). 
For example, if there are only two sensitive attributes (race and gender), it can be expressed as follows:
\begin{align*}
\mathcal{S}^{race}  \,\,\, = \,\,\, \{& Blue, Green, Purple \} \\
\mathcal{S}^{gender} \,\,\, = \,\,\, \{& male, female \} \\
S \,\,\, \in \,\,\,  \{& (Blue, male), (Green, male), \\
& (Purple, male), (Blue, female), \\
& (Green, female), (Purple, female) \}.
\end{align*}
We also define {$S$} as a set of subgroups.
The tuples for a subgroup are expressed as {$ \{s_1\ldots s_r \} $} for convenience.

We assume that the dataset consists of {$n$} instances {$ X=\{X_1,\ldots ,\\X_n\} $}.
The elements of {$X$} are tuples of {$S$}, {$A$}, {$C$}, and {$Z$}. 
Thus, each instance tuple is expressed as follows: 
\[
X_i \,\,\, = \,\,\, (S_i, A_i, C_i, Z_i).
\]
\noindent
Note that {$A$} is not used in describing our methods and definitions.

To apply the conventional fairness metrics to the conditions where there are multiple subgroups related to the sensitive attributes, we define a fair situation as the situation where the fairness metrics of all subgroups are equal.
This definition is different from the conventional definition of fairness to compare a protected group with another non-protected group.
In other words, we define the fair situation as the situation where the value of metrics for any subgroup is equal to that for all individuals as follows:
\begin{dfn}[Concept of subgroup fairness criteria]
\[
p(X, s) \,\,\, = \,\,\, p(X), \quad \forall{s}\in S.
\]
\label{dfn:fc}
\end{dfn}
While {$ p (X,s) $} calculates specific metrics for specified subgroup {$s$}, {$ p(X) $} calculates the concerned metrics for the dataset as a whole, i.e., all instances. 
For example, in a decision on a loan application, when {$p$} indicates the acceptance rate, if the acceptance rate for all customers is 50\%, {$p(X)=0.5$}.
In this example, when there are three subgroups {$a$}, {$b$}, and {$c$} on the sensitive attributes of the customers, and these three subgroups have the different acceptance rates: {$p(X,a)=0.3$}, {$p(X,b)=0.5$}, and {$p(X,c)=0.6$} respectively, this situation is unfair.
In contrast, if all three subgroups have the same acceptance rates: {$p(X,a)=0.5$}, {$p(X,b)=0.5$}, and {$p(X,c)=0.5$}, it is fair because {$p(X) = p(X,a)$}, {$= p(X,b)$}, and = {$p(X,c)$} are satisfied.

Based on this, we define the following three well-known fairness criteria for the situations with multiple protected groups. 
In the following definitions, the word ``subgroup'' refers to the subgroups related to sensitive attributes, such as the elements of $S$.

\begin{dfn}[Demographic Parity]{A classifier satisfies this criterion if the probability with which an instance in any subgroup is categorized into the favorable class is equal to that in the whole dataset:} 
\begin{eqnarray*}
P \left( Z=+ \mid S=s \right) \,\,\, = \,\,\, P \left( Z=+ \right), \quad
\forall{s}\in S.
\end{eqnarray*}
\label{dfn:dp}
\end{dfn}
\noindent
This definition is different from the general definition of demographic parity in that we do not consider a non-protected group.
Because we treat all subgroups equally, we adopt a definition of comparing the values of metrics among all subgroups with that of the whole data, rather than between that of protected and non-protected groups.

\begin{dfn}[Equalized Odds]{
A classifier satisfies this criterion if the probabilities with which an instance in any subgroup is rightly categorized into the favorable class (\textbf{True Positive Rate}, \textbf{TPR}), and wrongly categorized into the favorable class (\textbf{False Positive Rate}, \textbf{FPR}) are equal to those in the whole dataset:}
\begin{eqnarray*}
&&P \left( Z=+ \mid C=c, S=s \right) \\
&=& P \left( Z=+ \mid C=c \right),  \quad
\forall{c} \in \{+, -\}, \forall{s}\in S.
\end{eqnarray*}
\label{dfn:eodds}
\end{dfn}

\begin{dfn}[Equal Opportunity]{A classifier satisfies this criterion if the probability with which an instance in any subgroup is rightly categorized into the favorable class (\textbf{TPR}) is equal to those in the whole dataset:
}
\begin{eqnarray*}
&& P \left( Z=+ \mid C=+, S=s \right) \\
&=&P \left( Z=+ \mid C=+ \right),  \quad
\forall{s}\in S.
\end{eqnarray*}
\label{dfn:eopp}
\end{dfn}
\noindent
We do not set non-protected groups in the above two definitions,  either.
This is also because we treat all subgroups equally and do not set specific subgroups as protected groups on the basis of Definition~\ref{dfn:fc}.
For equalized odds, there are two terms of true positive rate (TPR) and false positive rate (FPR), and we adopt those mean values as the result: {$({\rm TPR} + {\rm FPR})/2$}.

\section{method}\label{section:method}

Our approach aims to mitigate intersectional bias while retaining the characteristics of conventional methods.
For this purpose, we propose a method to extend the conventional methods.

\subsection{Basic Idea of One-vs.-One Mitigation}

\begin{figure}[t]
  \begin{center}
    \includegraphics[width=\linewidth]{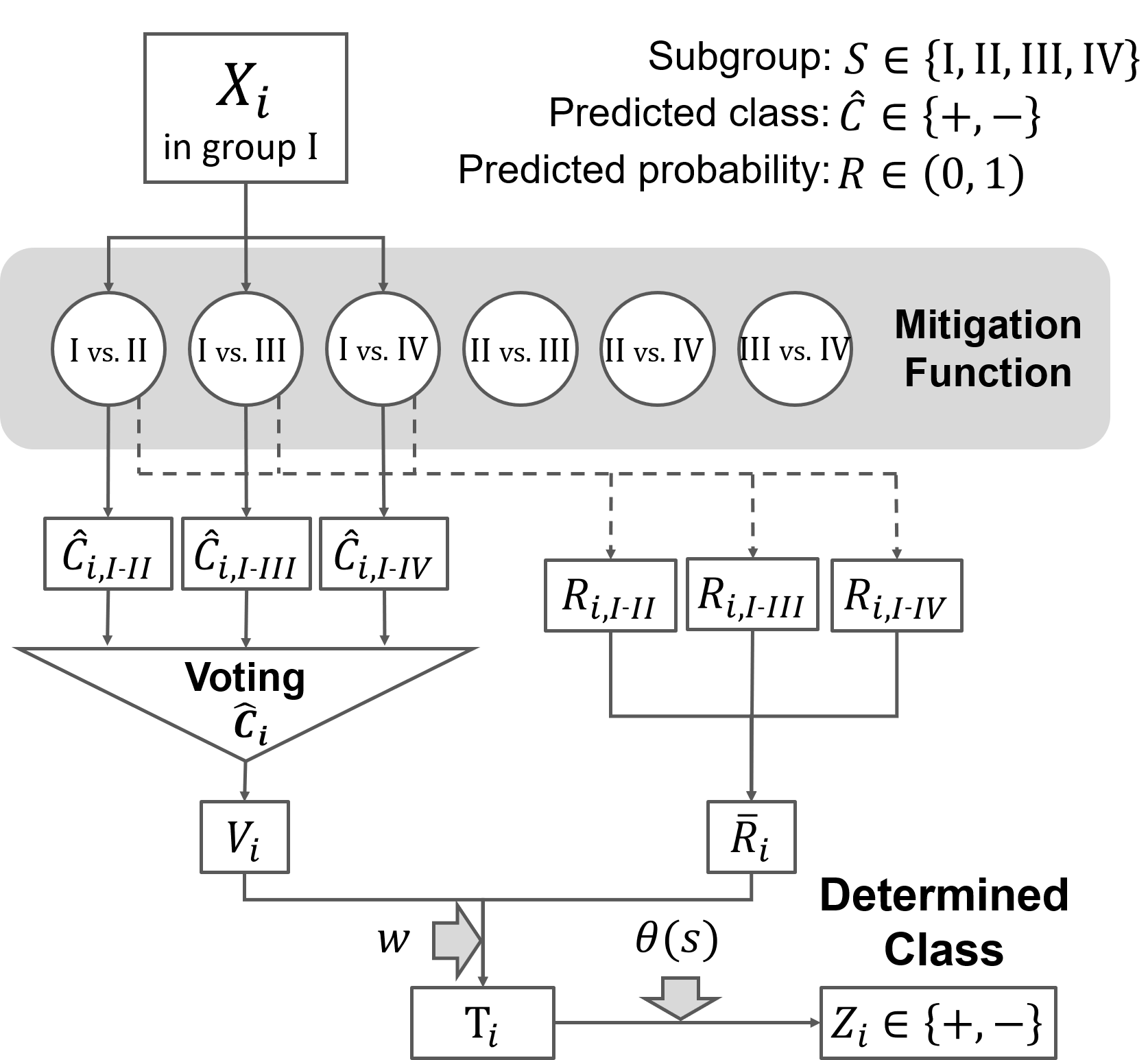}
    \caption{
    The overview diagram of our One-vs.-One Mitigation.
    Our method uses a bias mitigation function corresponding to each subgroup.
    When there are four subgroups {$ S=\{I,I\hspace{-.1em}I, I\hspace{-.1em}I\hspace{-.1em}I, I\hspace{-.1em}V\} $}, there are six subgroup pairs.
    In this example, we compare the results based on three subgroup pairs (I-I\hspace{-.1em}I, I-I\hspace{-.1em}I\hspace{-.1em}I, and I-I\hspace{-.1em}V) because $Xi$ belongs to I.
    Our method aggregates the mitigation results on each pairs, and calculates the $T_i$ based on a voting rate of the favorite class and the average value of the predicted probabilities.
    The final mitigation result $Zi$ is decided by whether the voting rate exceeds a threshold value $\theta(s)$.
    }
    \label{fig:ovo-m}
  \end{center}
\end{figure}

\begin{algorithm}[t]
\caption{One-vs.-One Mitigation}
\label{algo:ovom}
 \SetKwInOut{Input}{input}
 \SetKwInOut{Output}{output}
\Indm
\Input{{$ D $}: Original dataset, {$X_i$}: Test instance, \\${\bm{\theta}( = \{\theta(s)\})}$: Sets of score threshold for each subgroup,\\
{$\bm{S}$}: Sets of subgroups on sensitive attributes, \\ $M$: Mitigation function (model or method), \\
$w$: Trade-off parameter
}
\Output  {$Z_i$: Determined class}
\Indp

${\bf D^p} = \{D^p_1,\ldots ,D^p_J\} $ \\
${\bf D^p_{i}} = \{ D^p_j \in {\bf{D^p}}$ including $X_i$\}\\ 
${\bf \hat{C}_{i}} = {\bf \varnothing}$ \\
${\bf R_{i}} = {\bf \varnothing}$\\
\ForEach{$D^p_{i,j} \in {\bf D^p_{i}}$}{
$M: (X_i, D^p_{i,j}) \rightarrow (\hat{C}_{i, k}, R_{i,k})$\\
$\hat{C}_{i, k} \in \{ +, - \}$ add to ${\bf \hat{C}_{i}}$ \\ 
$R_{i, k} \in [0, 1]$ add to ${\bf R_{i}}$ 
}

$V_i \leftarrow $ ratio of $+$ in ${\bf \hat{C}_{i}}$\\
$\overline{R}_i \leftarrow $ average value in ${\bf R_{i}}$\\
$T_i = w * V_i + (1-w) * \overline{R}_i$

\eIf{$ T_i > \theta(S_i) $} {$ Z_i = + $} {$ Z_i = - $}
\end{algorithm}

We propose One-vs.-One Mitigation method that enables the general fairness-aware binary classification methods to mitigate the intersectional bias when there are multiple subgroups. 

As we summarized in Fig. \ref{fig:ovo-m} and described in Algorithm \ref{algo:ovom}, our method calculates the score for each instance using the majority vote results and predicted probability, which is obtained from the classification models or mitigation methods. 
First, our method divides the original dataset {$D$} into the sub-datasets {$ {\bf D^p} = \{D^p_1,\ldots ,D^p_J\} $} for each subgroup pair. {$J$} represents the number of subgroup pairs. 
For example, when there are four subgroups ({$I$}: (female, non-white), {${I\hspace{-.1em}I}$}: (female, white), {${I\hspace{-.1em}I\hspace{-.1em}I}$}: (male, non-white), {${I\hspace{-.1em}V}$}: (male, white)), there are six sub-datasets containing all possible subgroup pairs ($D^p_{I-I\hspace{-.1em}I}$, $D^p_{I-I\hspace{-.1em}I\hspace{-.1em}I}$, $D^p_{I-I\hspace{-.1em}V}$, $D^p_{I\hspace{-.1em}I-I\hspace{-.1em}I\hspace{-.1em}I}$, $D^p_{I\hspace{-.1em}I-I\hspace{-.1em}V}$, and $D^p_{I\hspace{-.1em}I\hspace{-.1em}I-I\hspace{-.1em}V}$).
For example, $D^p_{I-I\hspace{-.1em}I}$ includes all instances belonging to {$I$} and ones belonging to {${I\hspace{-.1em}I}$}.
Then, for each instance $X_i$, our method identifies ${\bf D^p_i}$, which is a subset of ${\bf D^p}$ that includes the subgroup to which $X_i$ belongs.
${\bf D^p_i}$ is identified to determine the subgroup pairs used when the subgroup that $X_i$ belongs is compared with the other subgroups.
With the above example, when an instance $X_i$ is categorized into {$I$}, ${\bf D^p_i}$ is identified as $\{D^p_{I-I\hspace{-.1em}I}, D^p_{I-I\hspace{-.1em}I\hspace{-.1em}I}, D^p_{I-I\hspace{-.1em}V}\}$.
Next, for each sub-dataset $D^p_{i, j} \in {\bf D^p_i}$, we obtain $\hat{C}_{i, k}$ and $R_{i, k}$ as the result of a mitigation function, letting {$K$} denote $|{\bf D^p_i}|$. Therefore, in the example we are taking, $K=3$. 
As the mitigation function, we use mitigated classification models trained with data for in-processing and post-processing methods, and a conventional mitigation method, which is not a classification model, for pre-processing methods.
This is because the training data processed by the pre-processing methods is modified before the models are built, and classification models cannot be prepared.
$\hat{C}_{i,k} \in \{+,-\}$ denotes the predicted class, and $R_{i,k} \in (0,1)$ denotes the predicted probability. 
When pre-processing methods that do not extract any predicted probability are used, $R_{i,k}$ is $0$.

Next, our method calculates $T_i$, which denotes the score for $X_i$, based on ${\bf \hat{C_{i}}} = \{\hat{C}_{i,1},\ldots,\hat{C}_{i,K}\}$ and ${\bf R_i} = \{R_{i,1},\ldots,R_{i,K}\}$.
Ideally, it is best to calculate the score $T_i$ based on only the results of the majority vote of $\hat{C}_{i,k}$ because we cannot obtain the information of the predicted probability from some mitigation methods such as reweighing~\cite{10.1007/s10115-011-0463-8} and disparate impact remover~\cite{10.1145/2783258.2783311}. 
Note that the result of the majority vote means the ratio of the favorable class out of all results obtained from the comparison between each pair of subgroups related to $X_i$. 
However, in our method, when there are only a small number of subgroups, if we use only the majority vote to determine the score, there are a lot of instances that have the same score values. 
This leads to that it is difficult to determine which instances should be selected to change their classes based on the score values.  
We then also use the predicted probability extracted from the classification methods to calculate the score. 
To balance the results of the majority vote and the predicted probability, we introduce  a trade-off parameter $w$. 
Using the above information, $T_i$ is defined as follows: 
\begin{eqnarray*}
    T_i = w*V_i + (1-w)*\overline{R}_i.
\end{eqnarray*}
\noindent
$V_i$ denotes the results of the majority vote, and $\overline{R}_i$ is the average value in ${\bf R_i}$. $w$ is a value set as $(|S|-1)/|S|$  by solving the equation of $w/(|S|-1) = 1-w$. The left-hand side of this equation is the value of score that changes when a vote increases or decreases, and the right-hand side is the upper bound  of $(1-w)*\overline{R}_i$, where $|S|$ denotes the number of subgroups.
This value of $w$ is set to prevent the effect of the predicted probability surpassing that of the results of the majority vote, and to make $T_i$ a continuous value.
In the example we are taking, $|S| = 4$ and $w = 3/4 = 0.75$. When $(\hat{C}_{i,I-I\hspace{-.1em}I}, \hat{C}_{i,I-I\hspace{-.1em}I\hspace{-.1em}I}, \hat{C}_{i,I-I\hspace{-.1em}V}) = (+, +, -)$, and $(R_{i,I-I\hspace{-.1em}I}, R_{i,I-I\hspace{-.1em}I\hspace{-.1em}I}, R_{i,I-I\hspace{-.1em}V}) = (0.8, 0.6, 0.4)$, $V_i = 2/3 =0.67$, and $\overline{R}_i = 1.8 / 3 = 0.6$. 
Therefore, nani$T_i = 0.75*0.67 + 0.25*0.6 = 0.65$.

Based on $T_i$, we identify the exact results that satisfy the appropriate value of the fairness metrics and accuracy.
To do this, we introduce $\theta (s)$, which denotes the threshold of the score used to determine the instances classified into the favorable class. 
If $T_i > \theta (s)$, $X_i$ is categorized into the favorable class, the determined class $Z_i = +$. Otherwise, $Z_i = -$.
A different $\theta(s)$ is set for a different subgroup because, in the situations with discrimination, it is necessary to make low-score instances in the strongly discriminated subgroup classified into the favorable class, and make high-score instances in the strongly privileged subgroup classified into the unfavorable class by setting a different threshold for each subgroup.

When setting $\theta(s)$, we attempt to find the point where the fairness is compatible with accuracy.
To control the trade-off between accuracy and fairness, users of our method can set an upper limit of disparity {$\epsilon$}.
Let {$ Q\bigl(T, \Theta \bigr)$} be an accuracy metric with a score {$ T $} and a set of score threshold {$\Theta$}. 
{$\Theta$} represents a set of {$\theta(s)$} for subgroup {$s$}.
Letting $T(s)$ denote a set of $T_i$ that belongs to subgroup $s$, $T$ represents a set of $T(s)$ for all subgroups.
And we introduce $\gamma$ to denote the disparity that means the extent to which the unfairness is included. 
We will describe how the values of $\gamma$ are calculated in our experiment in Section~\ref{sec:experiment}.
Our method searches for the {$\theta(s) \in \Theta$} that maximizes {$ Q\bigl(T, \Theta \bigr)$} within {$ \gamma < \epsilon $} at training time:
\begin{eqnarray*}
    \argmax_{\Theta \in [0,1]} \Bigl[ Q\bigl( T, \Theta \bigr) \mid \gamma < \epsilon \Bigr].
\end{eqnarray*}
\noindent
By increasing the value of {$\epsilon$}, the number of candidates {$\theta$} increases, and the value of the accuracy metric tends to improve.

\subsection{Application to Each Approach}
Our method is applied to conventional fairness-aware machine learning methods in different manners depending on its approach, i.e., pre-, in-, and post-processing. 
We describe how our method is applied to each approach.

\subsubsection{Preparation}
Before using our One-vs.-One Mitigation method, we split the whole dataset into the training dataset $U$ and test dataset $V$.

\subsubsection{Pre-processing}
For the pre-processing methods, our method is applied when the bias in the training data is mitigated.
Then, $U$ is used as $D$ and the chosen conventional method is used as $M$ in Algorithm~\ref{algo:ovom}.
As a result of Algorithm~\ref{algo:ovom}, the determined class $Z_i$ for each instance from the mitigated result is obtained. 
Based on the result, the mitigated dataset $\hat{U}$ is consisted. Therefore, in this approach, $C_i = Z_i$ in $\hat{U}$.
The mitigated dataset $\hat{U}$ is the final result of our method.

In our experiment, we built a plain classification model with a plain classifier (in our experiment, logistic regression) using the mitigated dataset $\hat{U}$. 
To make the condition of the experiment the same, we use the results extracted from the plain model to measure the fairness and accuracy metrics.

\subsubsection{In-processing}
For the in-processing methods, our method is applied at the prediction time, not at training time. 
For this approach, the mitigated models are built for all sub-dataset. 
In the example we are taking, six models are built. 

First, $U$ is separated into sub-dataset $U^p_j$ for each subgroup pair.
With each sub-dataset $U^p_j$, we build the prediction model with the chosen in-processing method.
After building the models for all subgroup pairs, our method is applied. 
$V$ is used as $D$, and each model is used as $M$ in Algorithm~\ref{algo:ovom}.
In this approach, $M$ differs in accordance with the subgroup to which $X_i$ belongs. 
In the example above, therefore, when $X_i$ belongs to $I$, three models built based on the pairs of $I-{I\hspace{-.1em}I}$, $I-{I\hspace{-.1em}I\hspace{-.1em}I}$, and $I-{I\hspace{-.1em}V}$ are used as $M$.

\subsubsection{Post-processing}
For the post-processing methods, our method is applied at the prediction time.
They generally use a plain classification model trained with a plain classifier first, and next, the mitigation model trained with the mitigation methods.
In the process we take, first, we built one prediction model with a plain classifier using the whole training dataset $U$.
After that, we divide $U$ into sub-dataset $U^p_j$ for each subgroup pair.
Then, we train the mitigation models with the sub-dataset $U^p_j$ for each subgroup pair.
Finally, we obtain one plain classification model and the same number of mitigation models as that of the subgroup pairs. 
In Algorithm~\ref{algo:ovom}, we use $V$ as $D$ and the combination of the normal and the mitigation models as $M$.
Therefore, in this approach, $M$ is different according to the subgroup which $X_i$ belongs to.
In the example above, when $X_i$ belongs to $I$, one plain classification model, and three models built based on the pairs of $I-{I\hspace{-.1em}I}$, $I-{I\hspace{-.1em}I\hspace{-.1em}I}$, and $I-{I\hspace{-.1em}V}$ are used as $M$.

\section{Experiment}
\label{sec:experiment}

In this section, we describe the settings of our experiment to measure the effect of intersectional bias mitigation of our method

\subsection{Methods and Fairness Criteria}

\begin{table}[]
\caption{Types and corresponding fairness criteria for method. 
}
\begin{tabular}{|l|c|c|c|c|}
\hline
\multicolumn{1}{|c|}{Method}   & Type & 
\begin{tabular}[c]{@{}c@{}}demographic\\parity\end{tabular} &
\begin{tabular}[c]{@{}c@{}}equalized\\odds\end{tabular} &
\begin{tabular}[c]{@{}c@{}}equal\\opportunity\end{tabular} \\ \hline
{\bf MS} \cite{10.1007/s10115-011-0463-8}           & Pre  &  \cm  & \xm  & \xm  \\ \hline
{\bf AD} \cite{10.1145/3278721.3278779}              & In   &  \cm  & \cm  & \cm \\ \hline {\bf ROC} \cite{6413831} & Post & \cm  & \xm & \xm \\ \hline 
{\bf EO} \cite{NIPS2016_6374}              & Post & \xm  & \cm & \cm \\ \hline 

\end{tabular}
\label{tb:methods}
\end{table}

We conduct experiments to show that our method can handle as diverse approaches and fairness criteria as possible.
We choose four methods and three criteria to show this as shown in Table \ref{tb:methods}. 
These methods are well-known as general ones for each approach of pre-, in-, and post-processing. By applying our method to them, we can show that our method works in diverse use-case scenarios.
Table \ref{tb:methods} shows the three types of methods and the correspondence between conventional bias mitigation methods and fairness criteria.
For the fairness criteria, we use demographic parity, equalized odds, and equal opportunity. 
Regarding pre-processing, only demographic parity is measured. 
On the other hand, regarding in- and post-processing, all three types of fairness criteria are measured. 
For post-processing, however, we use multiple methods to cover their criteria, because any one individual method does not support all fairness metrics. 

We briefly describe methods that we use as follows:

\begin{itemize}
\item {\bf MS} is {\it Massaging} developed in \cite{10.1007/s10115-011-0463-8} as {\bf pre-processing}. 
This method selects promotion and demotion (modifying favorable class labels to unfavorable and vice versa) instances for the training data in accordance with the predicted score by the classifier for pre-processing. 
We measure {\bf demographic parity} using this method.

\item {\bf AD} is {\it Adversarial Debiasing} developed in \cite{10.1145/3278721.3278779} as {\bf in-processing}, which aims to minimize the possibility to predict values of sensitive attributes from predicted class. 
We measure {\bf demographic parity}, {\bf equalized odds}, and {\bf equal opportunity} using this method.

\item {\bf ROC} is {\it Reject Option-based Classification} developed in \cite{6413831} as {\bf post-processing}, which modifies class labels near the classification threshold. 
At this time, instances in a protected group are modified to favorable class, and ones in a non-protected group are modified to unfavorable class.
We measure {\bf demographic parity} using this method. 

\item {\bf EO} is {\it Optimal Equalized Odds/Opportunity Predictor} developed in \cite{NIPS2016_6374} as {\bf post-processing}, which ensures no difference in true and false positive rates for prediction results between groups. 
This technology supports {\bf equalized odds} and {\bf equal opportunity}.

\end{itemize}

In our experiment, we compare the above conventional methods with the same methods to which our One-vs.-One Mitigation method  is applied, and with the plain classifier (logistic regression). 
When using the conventional methods, we adopt logistic regression as a general classifier used in pre- and post-processing.
As the results from the conventional methods, we prepare a different mitigation results on each sensitive attribute, not on all sensitive attributes. In other words, two mitigation results are prepared for each conventional method.
We do not use the mitigation results of the conventional methods based on multiple sensitive attributes because there are not any stable ways of mitigating bias on multiple sensitive attributes with the conventional methods.
For example, when the method mitigates bias on gender first and mitigates bias on race second, the effect of mitigation on gender will be eliminated.

\begin{table*}[htbp]
\small
\caption{Subgroup statistics for each dataset: (a)Adult, (b)COMPAS. Both dataset have same two sensitive attributes(race, gender), and four subgroups. These tables show the number of instances and their percentage of the overall for each subgroup.} 
\begin{tabular}{cc}
\hspace{-8mm}
\begin{minipage}[t]{0.5\hsize}
    \centering
    \subcaption{Adult}
    \begin{tabular}{|l|p{13.5mm}|p{13mm}|p{13mm}|p{14mm}|}
    \hline
    \multicolumn{2}{|l|}{\multirow{2}{*}{}} & \multicolumn{3}{l|}{gender} \\ \cline{3-5} 
    \multicolumn{2}{|l|}{}                  & male   & female  & overall  \\ \hline
    \multirow{3}{*}{race}    & white        & 27,020\,(60\%)     & 11,883\,(26\%)  & 38,903\,(86\%)       \\ \cline{2-5} 
                         & non-white    & 3,507\,(8\%)     & 2,812\,(6\%)  & 6,319\,(14\%)       \\ \cline{2-5} 
                         & overall      & 30,527\,(68\%)     & 14,695\,(32\%)  & 45,222\,(100\%)      \\ \hline
    \end{tabular}
\end{minipage} 
    
\begin{minipage}[t]{0.5\hsize}
    \centering
    \subcaption{COMPAS}
    \hspace{8mm}
    \begin{tabular}{|l|p{13.5mm}|p{13mm}|p{13mm}|p{14mm}|}
    \hline
    \multicolumn{2}{|l|}{\multirow{2}{*}{}} & \multicolumn{3}{l|}{gender} \\ \cline{3-5} 
    \multicolumn{2}{|l|}{}                  & male   & female  & overall  \\ \hline
    \multirow{3}{*}{race}    & white        & 1,620\,(26\%)     & 480\,(8\%)  & 2,100\,(34\%)       \\ \cline{2-5} 
                         & non-white    & 3,374\,(55\%)     & 693\,(11\%)  & 4,067\,(66\%)       \\ \cline{2-5} 
                         & overall      & 4,994\,(81\%)     & 1,173\,(19\%)  & 6,167\,(100\%)      \\ \hline
\end{tabular}
\end{minipage}  
\end{tabular}
\label{tb:dataset-stat}
\end{table*}

\subsection{Measurements}

To compare the performance of our method with the conventional methods and the plain classifier, we measure disparity and accuracy. 
The disparity is the value indicating how unfair the situation is.
We use 5-fold cross-validation to obtain results for all of the measurements.
Here, we set the upper limit of disparity $\epsilon=0.03$ to prioritize disparity reduction compared to guaranteeing the accuracy.

To set fairness criteria, we consider the definitions of disparity that are expressed as either a difference or a ratio.
Any metric can be used as both a difference or a ratio to measure the disparity.
For example, demographic parity can be used as a difference called statistical parity difference~\cite{10.1007/s10618-010-0190-x}, and a ratio called disparate impact~\cite{equal1990uniform}.
Therefore, considering both disparities, we define the disparity as the difference {$\gamma_d$} and as the ratio {$\gamma_r$} based on Definition \ref{dfn:fc} as follows:
\begin{dfn}[Subgroup Disparity]{}
\begin{eqnarray*}
{\gamma}_d &\coloneqq& \underset{{s}\in S}{\max} \Bigl|p(X) - p ( X, s ) \Bigr| ,\\
{\gamma}_r &\coloneqq& \underset{{s}\in S}{\max} \left \{ 1 - \min \left[ \dfrac{p ( X, s ) }{p(X)}, \dfrac{p(X)}{p ( X,s ) } \right] \right\} .\\
\end{eqnarray*}
\label{dfn:disparity}
\end{dfn}
\noindent
{$\gamma_d$} means the maximum absolute value of the difference between the value of a fairness metric of each subgroup and that of the whole data. 
The range of {$\gamma_d$} is {$ [0, 1] $}, and 0 is the ideal value. 
{$\gamma_r$} means the maximum value of one minus the minimum value of the ratio between the value of a fairness metric of each subgroup and that of the whole data. 
These disparities are defined to set their domain of definition as $[0, 1]$ and  their ideal value as $0$ to make it easy to compare among different combinations of methods and metrics. In this experiment, we measure {$\gamma_d$} and {$\gamma_r$} using Definitions \ref{dfn:dp}, \ref{dfn:eodds}, and \ref{dfn:eopp}.

As a criterion of accuracy, we measure balanced accuracy~\cite{balanced_accuracy}, which is applicable even for a dataset with an imbalance in the ratio of the positive and negative class. The balanced accuracy is defined as the mean of the true positive rate (TPR) and true negative rate (TNR), which is calculated as (TPR + TNR)/2.
Hereinafter, the balanced accuracy is referred to as accuracy.

After comparing our method with other methods, we investigate the effect of $\epsilon$ to the accuracy and disparity.
We change the value of {$\epsilon$} in the range of 0.03 to 0.99 at 0.33 intervals, and measure the disparity and accuracy in each $\epsilon$.

\subsection{Datasets}

We conducted our experiment on two real-world datasets: Adult~\cite{Dua:2017} and COMPAS~\cite{COMPASRe72:online}. 
Adult is a dataset of income of people based on the US Census. This dataset is used to predict whether a person's annual income is more than \$50k.  
The dataset has 45,222 instances with unknown values that are removed. 
COMPAS is a dataset used to predict recidivism within two years, and is well known that predictions using this dataset include a discriminatory bias \cite{propublica_machine_bias}. For the dataset, we use 6,167 instances using the same pre-processing as the original analysis\footnote{https://github.com/propublica/compas-analysis}, and removed unknown values.
Both datasets have race and gender as sensitive attributes. 
Adult and COMPAS have eleven and eight non-sensitive attributes, respectively. They are converted to one-hot vectors. 

Table \ref{tb:dataset-stat} shows the subgroup statistics for each dataset.
The sensitive attributes (race, gender) are binary, and there are four subgroups (i.e., non-white female, non-white male, white female, and white male).
For example, in Adult, it is shown that 38,903 white people account for 86\% of the overall number.

We divide the dataset into 4:1 for training and test data, respectively, per validation.

\subsection{Implementation}

We implement our proposed method using AIF360, an open-source toolkit of fairness-aware machine learning~\cite{DBLP:journals/corr/abs-1810-01943}. 
AIF360 includes many state-of-the-art bias mitigation methods and fairness criteria. 
We use this toolkit to support a wide range of bias mitigation methods and fairness criteria.
Additionally, we use the default values as hyperparameters of the conventional methods when we use the methods both with and without our method to compare the results in the same conditions.

\section{Results} \label{section:results}

In this section, we summarize the results of our experiment to compare the plain classifier (logistic regression), the conventional methods, and our method.

\def\gscale{0.5}
\begin{figure*}[t]
\begin{tabular}{cc}

    \begin{minipage}[t]{0.5\hsize}
    \begin{center}
        \hspace{-12mm}
        \includegraphics[scale=\gscale]{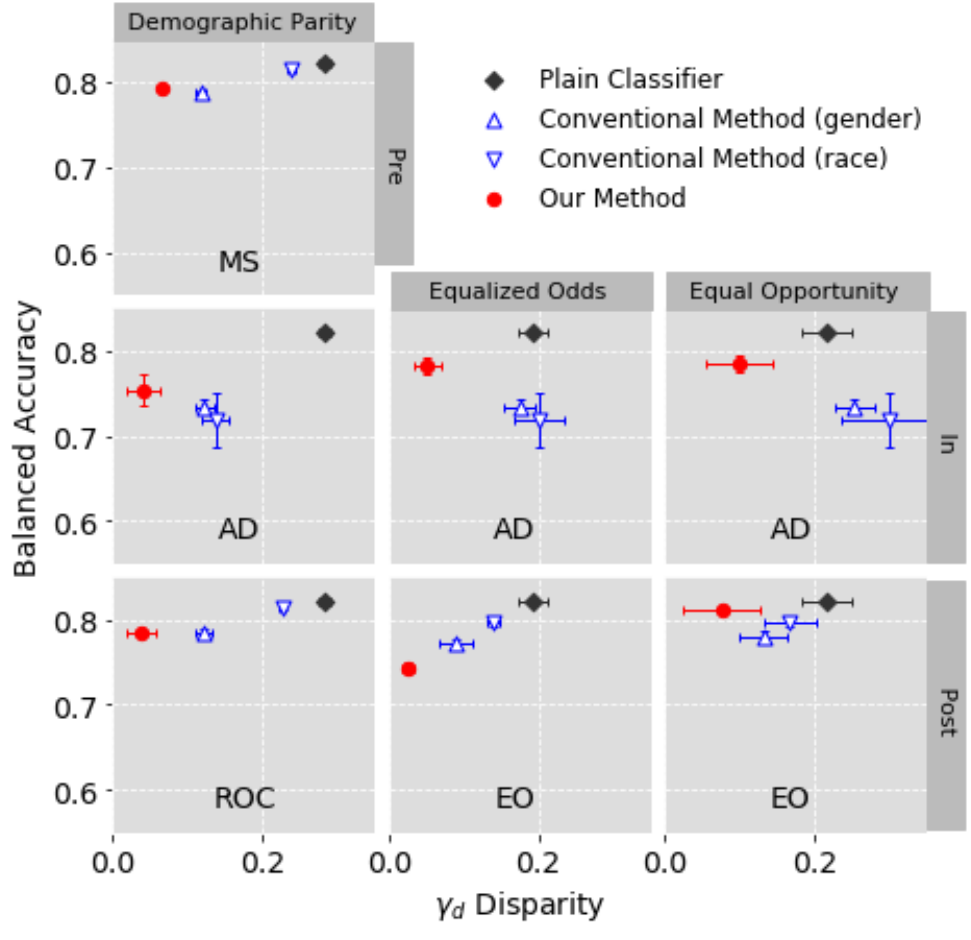}
        \subcaption{Disparity for difference on Adult}
        \label{fig:adult_diff}
    \end{center}
    \end{minipage}  
    
    \begin{minipage}[t]{0.5\hsize}
    \begin{center}
        \hspace{-7mm}
        \includegraphics[scale=\gscale]{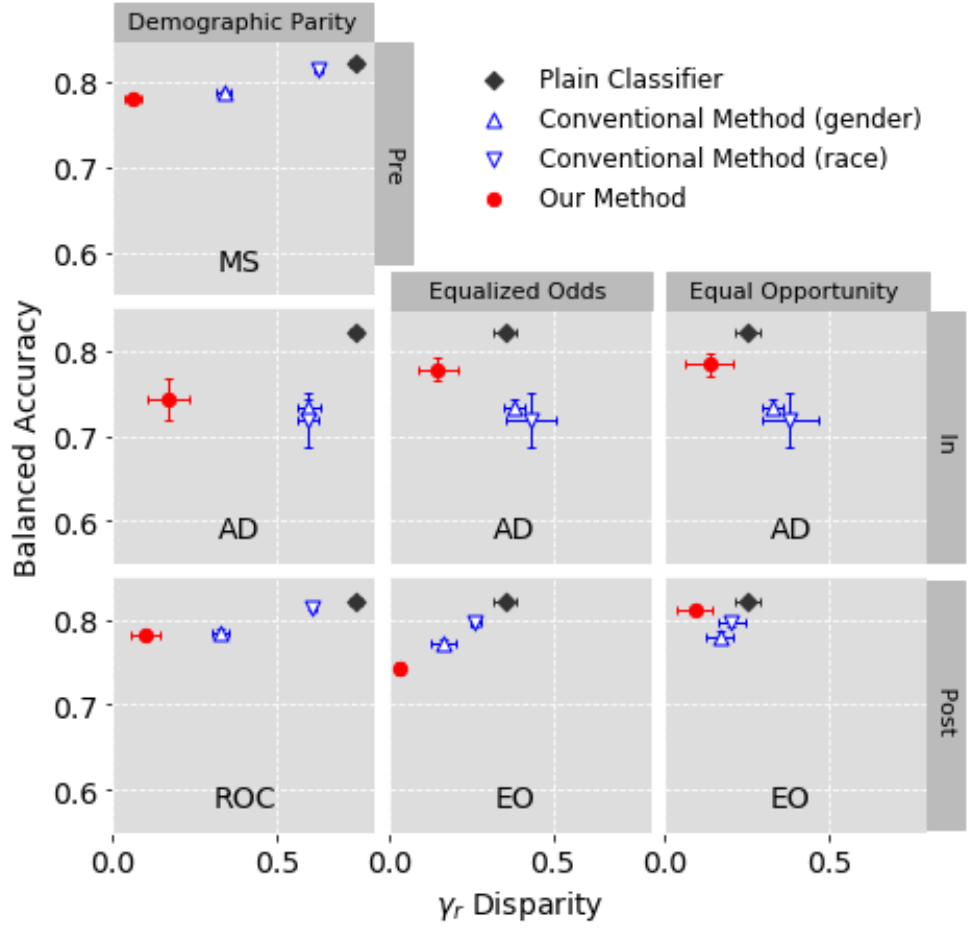}
        \subcaption{Disparity for ratio on Adult}
        \label{fig:adult_ratio}
    \end{center}
    \end{minipage} \\

    \begin{minipage}[t]{0.5\hsize}
    \begin{center}
        \vspace{2mm}
        \hspace{-12mm}
        \includegraphics[scale=\gscale]{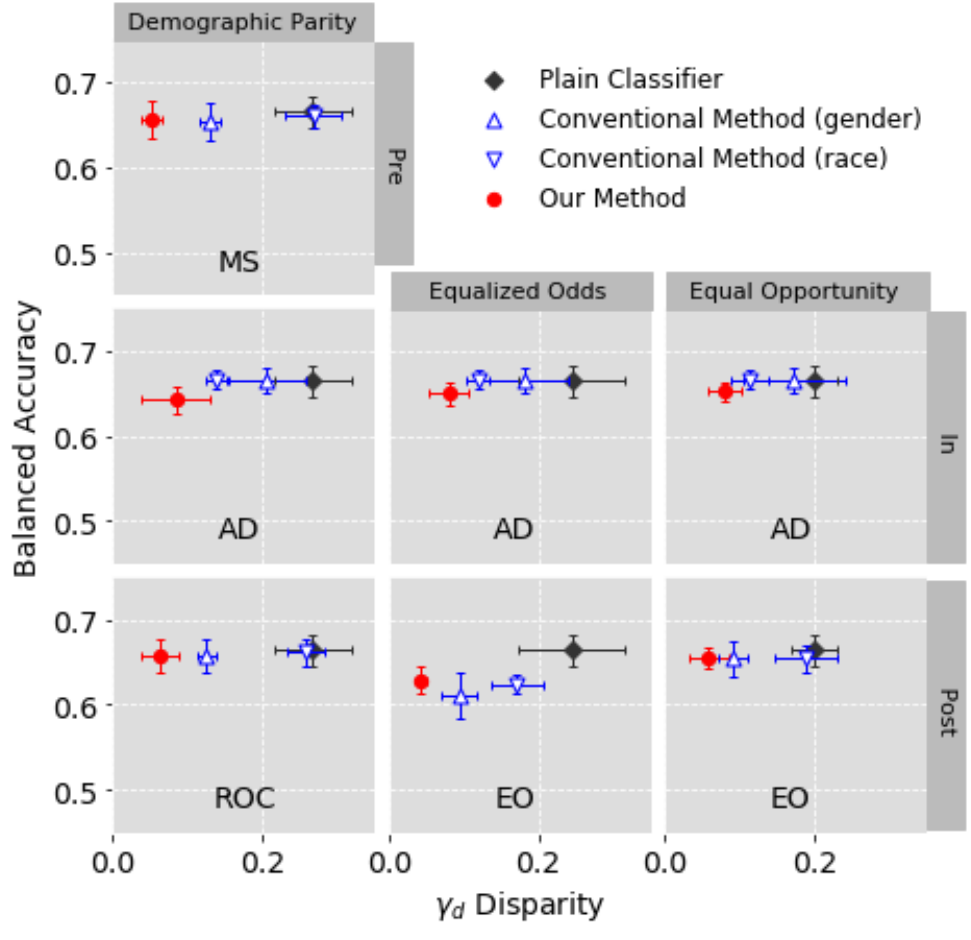}
        \subcaption{Disparity for difference on COMPAS}
        \label{fig:compas_diff}
    \end{center}
    \end{minipage}  
    
    \begin{minipage}[t]{0.5\hsize}
    \begin{center}
        \vspace{2mm}
        \hspace{-7mm}
        \includegraphics[scale=\gscale]{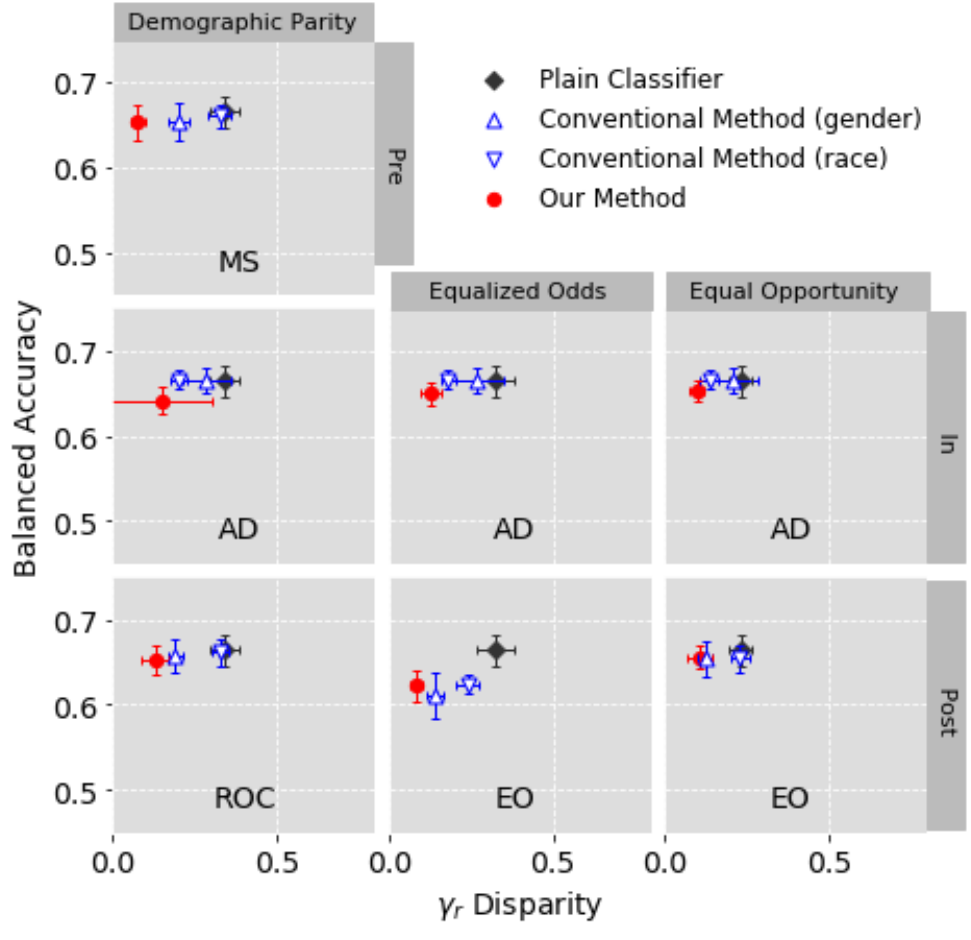}
        \subcaption{Disparity for ratio on COMPAS}
        \label{fig:compas_ratio}
    \end{center}
    \end{minipage} 
\end{tabular}
\caption{
Comparison of balanced accuracy and disparity for each method. (a)The disparity of {$\gamma_d$} on Adult. (b)The disparity of ratio {$\gamma_r$} on Adult. (c)The disparity of difference {$\gamma_d$} on COMPAS. (d)The disparity of ratio {$\gamma_r$} on COMPAS.
the X- and Y-axes represent disparity and balanced accuracy respectively.
Since the ideal value of the disparity is 0, and that of the balanced accuracy is 1, the further to the upper left a point is positioned, the better the result.
In each figure, the left column represents demographic parity, the middle column represents equalized odds, and the right column represents equal opportunity. 
The upper row represents pre-processing, the middle row represents in-processing, and the bottom row represents post-processing. 
All points and error bars represent the mean and standard deviations respectively, for 5-fold cross-validation. 
}
\label{fig:comparison_result}
\end{figure*}

\def\gscale{0.5}
\begin{figure*}[t]
\begin{tabular}{cc}

    \begin{minipage}[t]{0.5\hsize}
    \begin{center}
        \includegraphics[scale=\gscale]{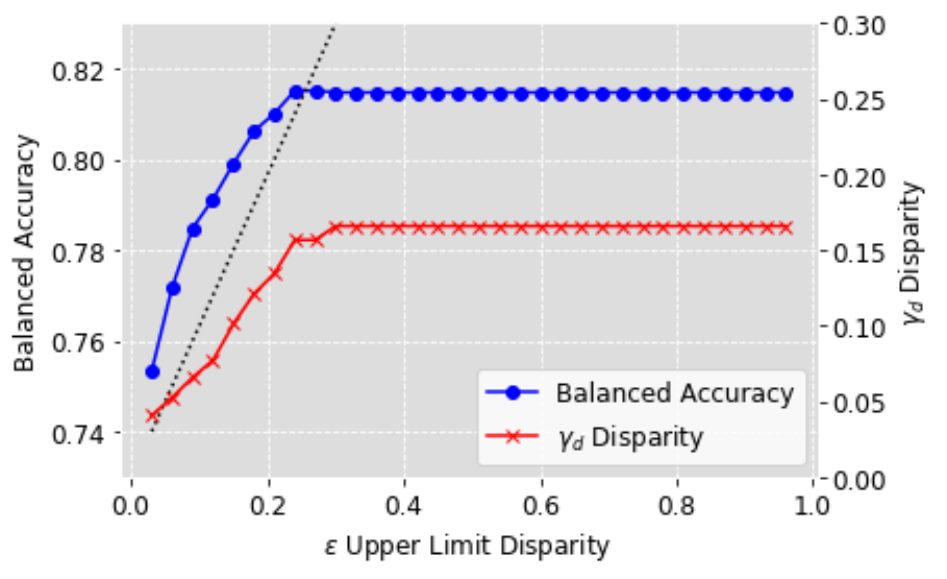}
        \subcaption{AD for demographic parity {$\gamma_d$} disparity}
        \label{fig:adult_AD_dp_diff_to}
    \end{center}
    \end{minipage}  
    
    \begin{minipage}[t]{0.5\hsize}
    \begin{center}
        \includegraphics[scale=\gscale]{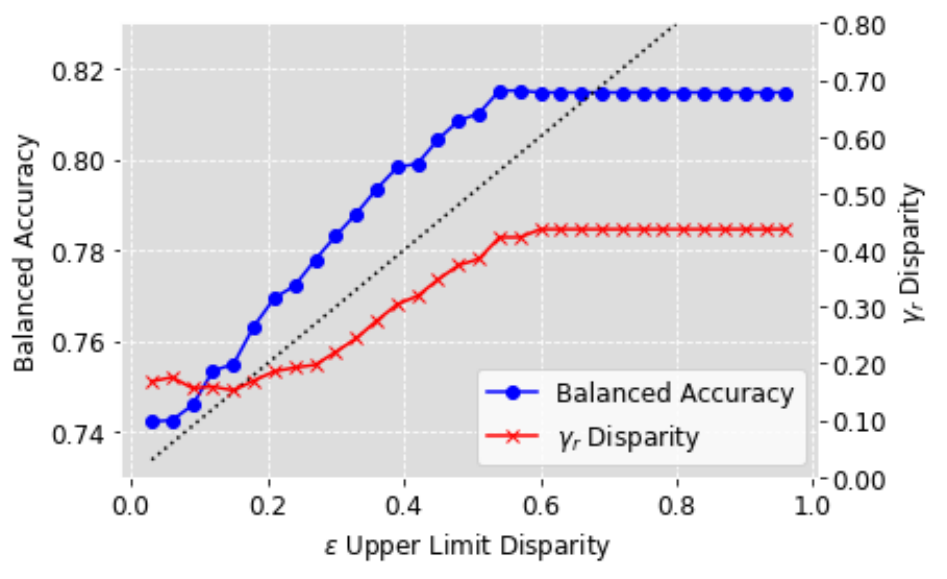}
        \subcaption{AD for demographic parity {$\gamma_r$} disparity}
        \label{fig:adult_AD_dp_ratio_to}
    \end{center}
    \end{minipage} \\

    \begin{minipage}[t]{0.5\hsize}
    \begin{center}
        \includegraphics[scale=\gscale]{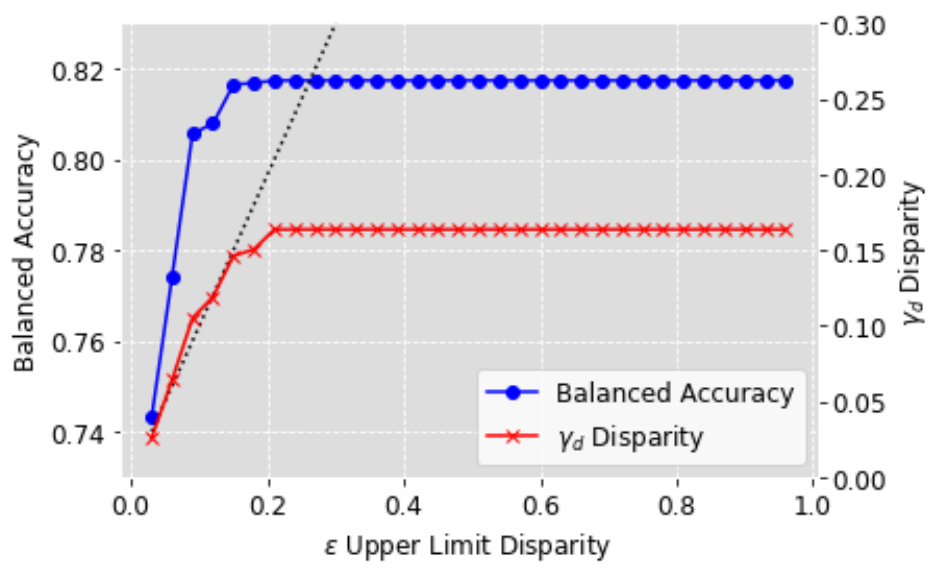}
        \subcaption{EO for equalized odds {$\gamma_d$} disparity}
        \label{fig:adult_EO_eodds_diff_to}
    \end{center}
    \end{minipage} 
    
    \begin{minipage}[t]{0.5\hsize}
    \begin{center}
        \includegraphics[scale=\gscale]{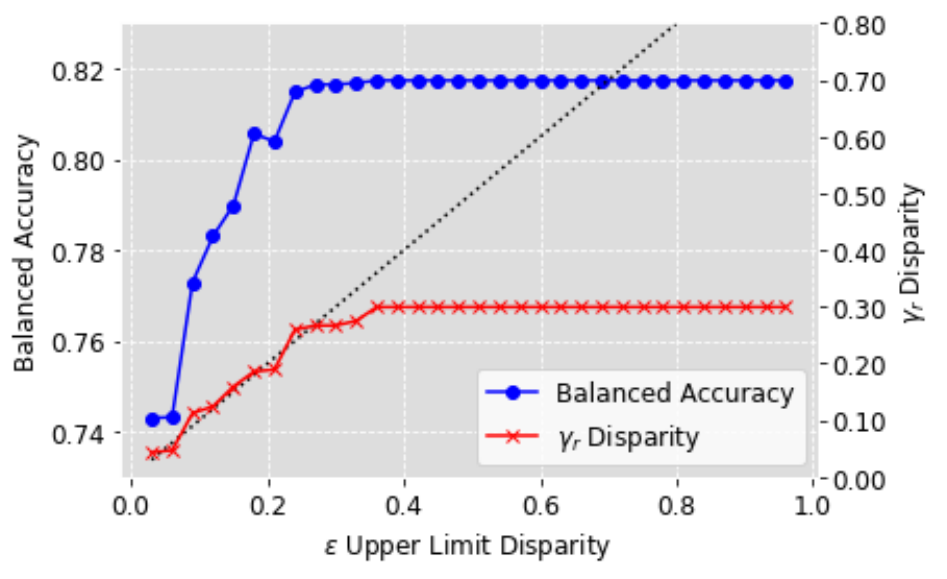}
        \subcaption{EO for equalized odds {$\gamma_r$} disparity}
        \label{fig:adult_EO_eodds_ratio_to}
    \end{center}
    \end{minipage} \\

\end{tabular}
\caption{Trade-off between balanced accuracy and {$\gamma$} with difference values of {$\epsilon$} on Adult dataset. 
The X-axis, left Y-axis, and right Y-axis represent {$\epsilon$}, balanced accuracy, and disparity, respectively. 
The dotted line represents disparity, which is measured with the right Y-axis, corresponding to the value of {$\epsilon$}, and, in the desired result, the disparity is lower than the line. 
Top row: Mitigation results for demographic parity by AD with our method. Bottom row: Mitigation results for equalized odds by EO with our method. The dotted line represents {$\epsilon$}, and it is desirable that the disparity is less than that line.}
\label{fig:adult_tradeoff}
\end{figure*}

\subsection{Comparison of Methods}\label{subsec:comparison_of_methods}

We summarize the results of the comparison of disparity and accuracy between our method and other methods as shown in Figure \ref{fig:comparison_result}.

\subsubsection{Adult Dataset}\label{subsection:comparison_methods_adult}

Figures \ref{fig:adult_diff} and \ref{fig:adult_ratio} show our experimental results on Adult. 
All of the results of our method show that disparities are reduced compared with other methods.
The disparities of difference improve to at most {$\gamma_d = 0.04 $} for demographic parity in {\bf ROC} , {$\gamma_d = 0.03 $} for equalized odds in {\bf EO}, and {$\gamma_d = 0.08 $} for equal opportunity in {\bf EO} in our method. 
Additionally, the disparities of ratio improve to at most {$\gamma_r = 0.06$} for demographic parity in {\bf MS}, {$\gamma_r = 0.04$} for equalized odds in {\bf EO}, and {$\gamma_r = 0.09$} for equal opportunity in {\bf EO} in our method.
For accuracy, compared to the conventional method, our method was worse in several methods and criteria. Especially, in \textbf{EO} for both the ratio and difference of equalized odds, the accuracy of our method (0.74) is clearly worse than the worse accuracy of the conventional method (gender, 0.77).  
On the other hand, compared to the conventional methods, our method significantly improved the disparities to {$(\gamma_d, \gamma_r)=(0.03,0.04)$} in \textbf{EO} for equalized odds.

Additionally, We observe that most conventional methods reduced both disparity and accuracy more than the plain classifier although, for {\bf AD} for equal opportunity, we also observe that the conventional methods increased the disparity compared with the plain classifier. 
On the other hand, in the plain classifier, there are large disparities with {$(\gamma_d, \gamma_r) = (0.28, 0.74) $} for demographic parity, {$(0.19, 0.35) $} for equalized odds, and {$(0.22, 0.25)$} for equal opportunity.

\subsubsection{COMPAS Dataset}

Figures \ref{fig:compas_diff} and \ref{fig:compas_ratio} show the results in the COMPAS dataset. 
As with Adult, our method reduced the disparity more than the conventional methods in all cases. The disparities of difference improves to at most {$\gamma_d = 0.05$} for demographic parity in {\bf MS}, {$\gamma_d = 0.04$} for equalized odds in {\bf EO}, and {$\gamma_d = 0.06$} for equal opportunity in {\bf EO} in our method.
The disparities of ratio improves to at most {$\gamma_r = 0.08$} for demographic parity in {\bf MS}, {$\gamma_r = 0.09$} for equalized odds in {\bf EO}, and {$\gamma_r = 0.10$} for equal opportunity in {\bf EO} in our method.
On the other hand, in regard to accuracy, there is no significant decline in the results of our method in all conditions from that of the conventional methods.

\subsection{Relationship among Accuracy, {$\gamma$} and {$\epsilon$}} \label{sec:acc_and_eps}

In this subsection, we report the effect of changes of upper limit disparity value {$\epsilon$} on accuracy and disparity. 
As described in Section \ref{section:method}, our method searches for a score threshold $\theta(s)$ that achieves the highest accuracy under the condition of {$\gamma < \epsilon$}. 
This can lead to the accuracy improving as the value of {$\epsilon$} increases. 
We pick {\bf AD} for demographic parity, and {\bf EO} for equalized odds on the Adult dataset as the combinations of methods and criteria to investigate the changes of accuracy and disparity. 
We choose them because they showed a clear decrease in accuracy in Figure \ref{fig:comparison_result}, and there is room to increase accuracy that helps us to investigate the effect of $\epsilon$ on the accuracy.

Figure \ref{fig:adult_tradeoff} shows the results of the changes disparity and accuracy when the value of {$\epsilon$} is changed. 
In Figure \ref{fig:adult_tradeoff}, we can see that as {$\epsilon$} increases within a certain value in any of the chosen combinations, both disparity and accuracy also increase. 
When {$\epsilon$} exceeded the certain value, both the accuracy and disparity converge. 
The accuracy after convergence was approximately a little less than 0.82, which is the value of accuracy of the plain classifier as shown in Figure \ref{fig:comparison_result} on Adult.

Figure \ref{fig:adult_AD_dp_diff_to} shows {\bf AD} for demographic parity by {$\gamma_d$}. 
If {$\epsilon$} = 0.03, then {$\gamma_d$} = 0.04, which is slightly above {$\epsilon$}, otherwise {$\gamma_d$} < {$\epsilon$} is satisfied. 
The result of {$\gamma_d = 0.17$} after convergence is better than the plain classifier disparity {$\gamma_d = 0.28 $} shown in Figure \ref{fig:adult_diff}. 
Figure \ref{fig:adult_AD_dp_ratio_to} shows {\bf AD} for demographic parity by {$\gamma_r$}. 
{$\gamma_r$} is larger than {$\epsilon$} in the range of {$\epsilon \leq 0.15$}, which is wider than the range of {$\epsilon \leq 0.03$} in the result of $\gamma_d$.
In the range of {$\epsilon \geq 0.18$}, our method satisfy {$ \gamma_r < \epsilon $}, and the value converges at {$ \gamma_r = 0.44 $}, which is also better than the plain classifier disparity for ratio {$ \gamma_r = 0.74 $} shown in Figure \ref{fig:adult_ratio}.

Figure \ref{fig:adult_EO_eodds_diff_to} shows {\bf EO} for equalized odds by {$\gamma_d$}. 
In Figure~ \ref{fig:adult_EO_eodds_diff_to}, {$\gamma_d$} follows along the dotted line of {$\epsilon$} as it increases before convergence. 
The value of {$\gamma_d = 0.16$} after convergence is better than the plain classifier disparity {$\gamma_d = 0.19 $} shown in Figure \ref{fig:adult_diff}.
Figure \ref{fig:adult_EO_eodds_ratio_to} shows {\bf EO} for equalized odds by {$\gamma_r$}. 
This {$\gamma_r$} also follows along the dotted line of {$\epsilon$} as it increases before convergence. 
Additionally, {$\gamma_r = 0.30$} after convergence, which is better than the plain classifier disparity {$\gamma_r = 0.35 $} shown in Figure \ref{fig:adult_diff}.

\section{Discussion} \label{section:discussion}

In this section, we discuss the effect of intersectional bias mitigation on disparity and accuracy based on the result of our experiment.

\subsection{Changes of Disparity}
\label{sec:tend_disp}
From the experimental results, we demonstrated that the disparity between subgroups decreases when the proposed One-vs.-One Mitigation was applied to the conventional methods. 
The results of our method were stable for all of the three approach types (pre-, in-, and post-processing).

Additionally, from the results in subsection~\ref{sec:acc_and_eps}, the disparities in the results of our method were better than that of the plain classifier after the values were converged.
This can be because our method uses the result of the majority vote of the predicted class by the conventional methods, and the mitigated results from the conventional methods are reflected in the final results of our methods.

In contrast, we show that most conventional methods reduced disparity worse than our method and their results were not stable from the perspective of intersectional bias.
The results of the conventional methods changed depending on the sensitive attribute they mitigated.
Additionally, it was even impossible to set a stable process to extract the results considering intersectional bias with the conventional methods.
In particular, in \textbf{AD} for equalized odds and equal opportunity, the conventional methods increased disparity rather than the plain classifier. 
One possible explanation for this is that we used a plain classifier whose accuracy was 0.82, which was higher than the accuracy of the plain logistic regression model used in \cite{10.1145/3278721.3278779}, 0.78. 
Because the values of equalized odds and equal opportunity are the metrics using true positive rate and false positive rate, they are affected by the accuracy of the model.
Therefore, it is possible that the disparities of the conventional method in \textbf{AD} were lower than that of the plain classifier probably because we used the more accurate model than the original paper of \textbf{AD}~\cite{10.1145/3278721.3278779}. 

Our results also implicated that it is more difficult to mitigate bias on the basis of the ratio of metrics ($\gamma_r$) than the difference ($\gamma_d$).
In Figure~\ref{fig:adult_AD_dp_diff_to} and \ref{fig:adult_AD_dp_ratio_to}, although the same conventional method (\textbf{AD}), dataset (Adult), and metrics (demographic parity) were used, $\gamma_r$  became lower than $\epsilon$ later than $\gamma_d$. 
This can be because, when the probability of each instance is categorized into favorable class  ({$p(X)$}) is small, $\gamma_r$ tends to become bigger more easily than $\gamma_d$ since the ratio of probability ($p(X,s)/p(X)$ ) tends to be affected by the value of $p(X)$ more than the difference of the probability ($p(X) - p(X,s)$).
Therefore, when $\epsilon$ is too small, even if the biggest $\theta (s)$ is set for the most privileged group, and the smallest $\theta (s)$ is set for the most unprivileged group, $\gamma_r$ does not satisfy $\epsilon$. 
This discussion implicates that when analysts use our method, they need to take the value of a metric for all instances ({$p(X)$}), and set appropriate {$\epsilon$} differently for {$\gamma_d$} and {$\gamma_r$}.

\subsection{Changes of Accuracy}
Our results showed that the loss of accuracy in our method compared to the conventional methods and the plain classifier unstably changes depending on the approach, metrics, and dataset applied.
In some results, e.g., the results of $\gamma_d$ disparity with the COMPAS dataset, there were almost no differences in accuracy among the results of our method, the conventional methods, and the plain classifier. There were even cases where our result achieved more accurate results than the conventional method (e.g., the results of $\gamma_d$ disparity of \textbf{EO} with COMPAS).
However, especially in the results of Adult dataset, the accuracy clearly decreased in some results of our method compared to the other methods and classifier (e.g., the result of $\gamma_r$ with Adult using \textbf{EO}).
These results may have occurred due to the difference of the balance of negative and positive classes between the datasets because while COMPAS has 3,358 positive and 2,809 negative class instances, Adult has 11,208 positive and 34,014 negative class instances. 
It is notable that the results of the accuracy affected by dataset. For further study, it may be effective to investigate more diverse accuracy metrics depending on the dataset used.

Additionally, from the results in subsection~\ref{sec:acc_and_eps}, we can confirm that the accuracy improved as {$\epsilon$}, which specifies the permissible disparity, increased.
Therefore, to achieve a result with a certain accuracy, it is helpful for the users of our method to select their appropriate $\epsilon$.

\section{conclusion} \label{section:conclusion}

In this paper, we proposed the One-vs.-One Mitigation method that enables fairness-aware binary classification methods to mitigate intersectional bias when there are multiple sensitive attributes.
Our method inherits the characteristics of three approaches of the conventional methods (pre-, in-, post-processing), and handles fairness criteria related to the disparate parity and the error rates.
With two real-world datasets, Adult and COMPAS, we demonstrated that our method mitigated intersectional bias better than the conventional methods and a plain classifier in all experimental results. 
We also confirmed that our method can control the trade-off between accuracy and fairness by adjusting the upper limit of disparity.
With these results,  we showed that our method is capable of mitigating intersectional bias in diverse real-world situations while satisfying a wide range of fairness requirements.
With our method, we expect that the burden of considering intersectional bias will be reduced when new methods of fairness-aware binary classification and criteria are proposed in the future.


\bibliographystyle{ACM-Reference-Format}
\bibliography{main}


\begin{thebibliography}{27}


\ifx \showCODEN    \undefined \def \showCODEN     #1{\unskip}     \fi
\ifx \showDOI      \undefined \def \showDOI       #1{#1}\fi
\ifx \showISBNx    \undefined \def \showISBNx     #1{\unskip}     \fi
\ifx \showISBNxiii \undefined \def \showISBNxiii  #1{\unskip}     \fi
\ifx \showISSN     \undefined \def \showISSN      #1{\unskip}     \fi
\ifx \showLCCN     \undefined \def \showLCCN      #1{\unskip}     \fi
\ifx \shownote     \undefined \def \shownote      #1{#1}          \fi
\ifx \showarticletitle \undefined \def \showarticletitle #1{#1}   \fi
\ifx \showURL      \undefined \def \showURL       {\relax}        \fi
\providecommand\bibfield[2]{#2}
\providecommand\bibinfo[2]{#2}
\providecommand\natexlab[1]{#1}
\providecommand\showeprint[2][]{arXiv:#2}

\bibitem[\protect\citeauthoryear{Angwin, Larson, Mattu, and Kirchner}{Angwin
  et~al\mbox{.}}{2016}]%
        {propublica_machine_bias}
\bibfield{author}{\bibinfo{person}{Julia Angwin}, \bibinfo{person}{Jeff
  Larson}, \bibinfo{person}{Surya Mattu}, {and} \bibinfo{person}{Lauren
  Kirchner}.} \bibinfo{year}{2016}\natexlab{}.
\newblock \showarticletitle{Machine Bias}.
\newblock \bibinfo{journal}{\emph{ProPublica (May 23 2016)}}
  (\bibinfo{year}{2016}).
\newblock
\urldef\tempurl%
\url{https://www.propublica.org/article/machine-bias-risk-assessments-in-criminal-sentencing}
\showURL{%
\tempurl}


\bibitem[\protect\citeauthoryear{Barocas and Selbst}{Barocas and
  Selbst}{2016}]%
        {Barocas-BigDataDI-2016}
\bibfield{author}{\bibinfo{person}{Solon Barocas} {and}
  \bibinfo{person}{Andrew~D. Selbst}.} \bibinfo{year}{2016}\natexlab{}.
\newblock \showarticletitle{Big Data's Disparate Impact}.
\newblock \bibinfo{journal}{\emph{California Law Review}}
  \bibinfo{volume}{104}, \bibinfo{number}{3} (\bibinfo{year}{2016}),
  \bibinfo{pages}{671--732}.
\newblock
\showISSN{00081221}
\urldef\tempurl%
\url{http://www.jstor.org/stable/24758720}
\showURL{%
\tempurl}


\bibitem[\protect\citeauthoryear{Bellamy, Dey, Hind, Hoffman, Houde, Kannan,
  Lohia, Martino, Mehta, Mojsilovic, Nagar, Ramamurthy, Richards, Saha,
  Sattigeri, Singh, Varshney, and Zhang}{Bellamy et~al\mbox{.}}{2018}]%
        {DBLP:journals/corr/abs-1810-01943}
\bibfield{author}{\bibinfo{person}{Rachel K.~E. Bellamy},
  \bibinfo{person}{Kuntal Dey}, \bibinfo{person}{Michael Hind},
  \bibinfo{person}{Samuel~C. Hoffman}, \bibinfo{person}{Stephanie Houde},
  \bibinfo{person}{Kalapriya Kannan}, \bibinfo{person}{Pranay Lohia},
  \bibinfo{person}{Jacquelyn Martino}, \bibinfo{person}{Sameep Mehta},
  \bibinfo{person}{Aleksandra Mojsilovic}, \bibinfo{person}{Seema Nagar},
  \bibinfo{person}{Karthikeyan~Natesan Ramamurthy}, \bibinfo{person}{John~T.
  Richards}, \bibinfo{person}{Diptikalyan Saha}, \bibinfo{person}{Prasanna
  Sattigeri}, \bibinfo{person}{Moninder Singh}, \bibinfo{person}{Kush~R.
  Varshney}, {and} \bibinfo{person}{Yunfeng Zhang}.}
  \bibinfo{year}{2018}\natexlab{}.
\newblock \showarticletitle{{AI} Fairness 360: An Extensible Toolkit for
  Detecting, Understanding, and Mitigating Unwanted Algorithmic Bias}.
\newblock \bibinfo{journal}{\emph{CoRR}}  \bibinfo{volume}{abs/1810.01943}
  (\bibinfo{year}{2018}).
\newblock
\showeprint[arxiv]{1810.01943}
\urldef\tempurl%
\url{http://arxiv.org/abs/1810.01943}
\showURL{%
\tempurl}


\bibitem[\protect\citeauthoryear{Brodersen, Ong, Stephan, and
  Buhmann}{Brodersen et~al\mbox{.}}{2010}]%
        {balanced_accuracy}
\bibfield{author}{\bibinfo{person}{Kay~Henning Brodersen},
  \bibinfo{person}{Cheng~Soon Ong}, \bibinfo{person}{Klaas~Enno Stephan}, {and}
  \bibinfo{person}{Joachim~M. Buhmann}.} \bibinfo{year}{2010}\natexlab{}.
\newblock \showarticletitle{The Balanced Accuracy and Its Posterior
  Distribution}. In \bibinfo{booktitle}{\emph{Proceedings of the 2010 20th
  International Conference on Pattern Recognition}}
  \emph{(\bibinfo{series}{ICPR '10})}. \bibinfo{publisher}{IEEE Computer
  Society}, \bibinfo{address}{USA}, \bibinfo{pages}{3121^^e2^^80^^933124}.
\newblock
\showISBNx{9780769541099}
\urldef\tempurl%
\url{https://doi.org/10.1109/ICPR.2010.764}
\showDOI{\tempurl}


\bibitem[\protect\citeauthoryear{Buolamwini and Gebru}{Buolamwini and
  Gebru}{2018}]%
        {gendershadespmlr-v81-buolamwini18a}
\bibfield{author}{\bibinfo{person}{Joy Buolamwini} {and}
  \bibinfo{person}{Timnit Gebru}.} \bibinfo{year}{2018}\natexlab{}.
\newblock \showarticletitle{Gender Shades: Intersectional Accuracy Disparities
  in Commercial Gender Classification}. In
  \bibinfo{booktitle}{\emph{Proceedings of the 1st Conference on Fairness,
  Accountability and Transparency}} \emph{(\bibinfo{series}{Proceedings of
  Machine Learning Research})}, \bibfield{editor}{\bibinfo{person}{Sorelle~A.
  Friedler} {and} \bibinfo{person}{Christo Wilson}} (Eds.),
  Vol.~\bibinfo{volume}{81}. \bibinfo{publisher}{PMLR}, \bibinfo{address}{New
  York, NY, USA}, \bibinfo{pages}{77--91}.
\newblock
\urldef\tempurl%
\url{http://proceedings.mlr.press/v81/buolamwini18a.html}
\showURL{%
\tempurl}


\bibitem[\protect\citeauthoryear{^^c3^^81. A.~{Cabrera}, {Epperson}, {Hohman},
  {Kahng}, {Morgenstern}, and {Chau}}{^^c3^^81. A.~{Cabrera}
  et~al\mbox{.}}{2019}]%
        {Cabrera-fairvis-2019}
\bibfield{author}{\bibinfo{person}{^^c3^^81. A.~{Cabrera}}, \bibinfo{person}{W.
  {Epperson}}, \bibinfo{person}{F. {Hohman}}, \bibinfo{person}{M. {Kahng}},
  \bibinfo{person}{J. {Morgenstern}}, {and} \bibinfo{person}{D.~H. {Chau}}.}
  \bibinfo{year}{2019}\natexlab{}.
\newblock \showarticletitle{FAIRVIS: Visual Analytics for Discovering
  Intersectional Bias in Machine Learning}. In \bibinfo{booktitle}{\emph{2019
  IEEE Conference on Visual Analytics Science and Technology (VAST)}}.
  \bibinfo{pages}{46--56}.
\newblock


\bibitem[\protect\citeauthoryear{Calders and Verwer}{Calders and
  Verwer}{2010}]%
        {10.1007/s10618-010-0190-x}
\bibfield{author}{\bibinfo{person}{Toon Calders} {and} \bibinfo{person}{Sicco
  Verwer}.} \bibinfo{year}{2010}\natexlab{}.
\newblock \showarticletitle{Three Naive Bayes Approaches for
  Discrimination-Free Classification}.
\newblock \bibinfo{journal}{\emph{Data Min. Knowl. Discov.}}
  \bibinfo{volume}{21}, \bibinfo{number}{2} (\bibinfo{date}{Sept.}
  \bibinfo{year}{2010}), \bibinfo{pages}{277^^e2^^80^^93292}.
\newblock
\showISSN{1384-5810}
\urldef\tempurl%
\url{https://doi.org/10.1007/s10618-010-0190-x}
\showDOI{\tempurl}


\bibitem[\protect\citeauthoryear{Calmon, Wei, Vinzamuri, Ramamurthy, and
  Varshney}{Calmon et~al\mbox{.}}{2017}]%
        {NIPS2017_6988}
\bibfield{author}{\bibinfo{person}{Flavio~P. Calmon}, \bibinfo{person}{Dennis
  Wei}, \bibinfo{person}{Bhanukiran Vinzamuri},
  \bibinfo{person}{Karthikeyan~Natesan Ramamurthy}, {and}
  \bibinfo{person}{Kush~R. Varshney}.} \bibinfo{year}{2017}\natexlab{}.
\newblock \showarticletitle{Optimized Pre-Processing for Discrimination
  Prevention}. In \bibinfo{booktitle}{\emph{Proceedings of the 31st
  International Conference on Neural Information Processing Systems}} (Long
  Beach, California, USA) \emph{(\bibinfo{series}{NIPS'17})}.
  \bibinfo{publisher}{Curran Associates Inc.}, \bibinfo{address}{Red Hook, NY,
  USA}, \bibinfo{pages}{3995^^e2^^80^^934004}.
\newblock
\showISBNx{9781510860964}


\bibitem[\protect\citeauthoryear{Celis, Huang, Keswani, and Vishnoi}{Celis
  et~al\mbox{.}}{2019}]%
        {10.1145/3287560.3287586}
\bibfield{author}{\bibinfo{person}{L.~Elisa Celis}, \bibinfo{person}{Lingxiao
  Huang}, \bibinfo{person}{Vijay Keswani}, {and} \bibinfo{person}{Nisheeth~K.
  Vishnoi}.} \bibinfo{year}{2019}\natexlab{}.
\newblock \showarticletitle{Classification with Fairness Constraints: A
  Meta-Algorithm with Provable Guarantees}. In
  \bibinfo{booktitle}{\emph{Proceedings of the Conference on Fairness,
  Accountability, and Transparency}} (Atlanta, GA, USA)
  \emph{(\bibinfo{series}{FAT* '19})}. \bibinfo{publisher}{Association for
  Computing Machinery}, \bibinfo{address}{New York, NY, USA},
  \bibinfo{pages}{319^^e2^^80^^93328}.
\newblock
\showISBNx{9781450361255}
\urldef\tempurl%
\url{https://doi.org/10.1145/3287560.3287586}
\showDOI{\tempurl}


\bibitem[\protect\citeauthoryear{Commission}{Commission}{1978}]%
        {equal1990uniform}
\bibfield{author}{\bibinfo{person}{Equal Employment~Opportunity Commission}.}
  \bibinfo{year}{1978}\natexlab{}.
\newblock \showarticletitle{Uniform guidelines on employee selection
  procedures}.
\newblock  (\bibinfo{year}{1978}).
\newblock


\bibitem[\protect\citeauthoryear{Crenshaw}{Crenshaw}{1989}]%
        {crenshaw1989demarginalizing}
\bibfield{author}{\bibinfo{person}{Kimberle Crenshaw}.}
  \bibinfo{year}{1989}\natexlab{}.
\newblock \showarticletitle{Demarginalizing the intersection of race and sex: A
  black feminist critique of antidiscrimination doctrine, feminist theory and
  antiracist politics}.
\newblock \bibinfo{journal}{\emph{u. Chi. Legal f.}} (\bibinfo{year}{1989}),
  \bibinfo{pages}{139}.
\newblock


\bibitem[\protect\citeauthoryear{Dheeru and Karra~Taniskidou}{Dheeru and
  Karra~Taniskidou}{2017}]%
        {Dua:2017}
\bibfield{author}{\bibinfo{person}{Dua Dheeru} {and} \bibinfo{person}{Efi
  Karra~Taniskidou}.} \bibinfo{year}{2017}\natexlab{}.
\newblock \bibinfo{title}{{UCI} Machine Learning Repository}.
\newblock
\newblock
\urldef\tempurl%
\url{http://archive.ics.uci.edu/ml}
\showURL{%
\tempurl}


\bibitem[\protect\citeauthoryear{Dwork, Hardt, Pitassi, Reingold, and
  Zemel}{Dwork et~al\mbox{.}}{2012}]%
        {Dwork-2012-Awareness}
\bibfield{author}{\bibinfo{person}{Cynthia Dwork}, \bibinfo{person}{Moritz
  Hardt}, \bibinfo{person}{Toniann Pitassi}, \bibinfo{person}{Omer Reingold},
  {and} \bibinfo{person}{Richard Zemel}.} \bibinfo{year}{2012}\natexlab{}.
\newblock \showarticletitle{Fairness through Awareness}. In
  \bibinfo{booktitle}{\emph{Proceedings of the 3rd Innovations in Theoretical
  Computer Science Conference}} (Cambridge, Massachusetts)
  \emph{(\bibinfo{series}{ITCS '12})}. \bibinfo{publisher}{Association for
  Computing Machinery}, \bibinfo{address}{New York, NY, USA},
  \bibinfo{pages}{214^^e2^^80^^93226}.
\newblock
\showISBNx{9781450311151}
\urldef\tempurl%
\url{https://doi.org/10.1145/2090236.2090255}
\showDOI{\tempurl}


\bibitem[\protect\citeauthoryear{Feldman, Friedler, Moeller, Scheidegger, and
  Venkatasubramanian}{Feldman et~al\mbox{.}}{2015}]%
        {10.1145/2783258.2783311}
\bibfield{author}{\bibinfo{person}{Michael Feldman},
  \bibinfo{person}{Sorelle~A. Friedler}, \bibinfo{person}{John Moeller},
  \bibinfo{person}{Carlos Scheidegger}, {and} \bibinfo{person}{Suresh
  Venkatasubramanian}.} \bibinfo{year}{2015}\natexlab{}.
\newblock \showarticletitle{Certifying and Removing Disparate Impact}. In
  \bibinfo{booktitle}{\emph{Proceedings of the 21th ACM SIGKDD International
  Conference on Knowledge Discovery and Data Mining}} (Sydney, NSW, Australia)
  \emph{(\bibinfo{series}{KDD '15})}. \bibinfo{publisher}{Association for
  Computing Machinery}, \bibinfo{address}{New York, NY, USA},
  \bibinfo{pages}{259^^e2^^80^^93268}.
\newblock
\showISBNx{9781450336642}
\urldef\tempurl%
\url{https://doi.org/10.1145/2783258.2783311}
\showDOI{\tempurl}


\bibitem[\protect\citeauthoryear{Foulds, Islam, Keya, and Pan}{Foulds
  et~al\mbox{.}}{2020}]%
        {foulds2020intersectional}
\bibfield{author}{\bibinfo{person}{James~R Foulds}, \bibinfo{person}{Rashidul
  Islam}, \bibinfo{person}{Kamrun~Naher Keya}, {and} \bibinfo{person}{Shimei
  Pan}.} \bibinfo{year}{2020}\natexlab{}.
\newblock \showarticletitle{An intersectional definition of fairness}. In
  \bibinfo{booktitle}{\emph{2020 IEEE 36th International Conference on Data
  Engineering (ICDE)}}. IEEE, \bibinfo{pages}{1918--1921}.
\newblock


\bibitem[\protect\citeauthoryear{Hardt, Price, and Srebro}{Hardt
  et~al\mbox{.}}{2016}]%
        {NIPS2016_6374}
\bibfield{author}{\bibinfo{person}{Moritz Hardt}, \bibinfo{person}{Eric Price},
  {and} \bibinfo{person}{Nathan Srebro}.} \bibinfo{year}{2016}\natexlab{}.
\newblock \showarticletitle{Equality of Opportunity in Supervised Learning}. In
  \bibinfo{booktitle}{\emph{Proceedings of the 30th International Conference on
  Neural Information Processing Systems}} (Barcelona, Spain)
  \emph{(\bibinfo{series}{NIPS'16})}. \bibinfo{publisher}{Curran Associates
  Inc.}, \bibinfo{address}{Red Hook, NY, USA},
  \bibinfo{pages}{3323^^e2^^80^^933331}.
\newblock
\showISBNx{9781510838819}


\bibitem[\protect\citeauthoryear{Hebert-Johnson, Kim, Reingold, and
  Rothblum}{Hebert-Johnson et~al\mbox{.}}{2018}]%
        {pmlr-v80-hebert-johnson18a}
\bibfield{author}{\bibinfo{person}{Ursula Hebert-Johnson},
  \bibinfo{person}{Michael Kim}, \bibinfo{person}{Omer Reingold}, {and}
  \bibinfo{person}{Guy Rothblum}.} \bibinfo{year}{2018}\natexlab{}.
\newblock \showarticletitle{Multicalibration: Calibration for the
  ({C}omputationally-Identifiable) Masses}. In
  \bibinfo{booktitle}{\emph{Proceedings of the 35th International Conference on
  Machine Learning}} \emph{(\bibinfo{series}{Proceedings of Machine Learning
  Research})}, \bibfield{editor}{\bibinfo{person}{Jennifer Dy} {and}
  \bibinfo{person}{Andreas Krause}} (Eds.), Vol.~\bibinfo{volume}{80}.
  \bibinfo{publisher}{PMLR}, \bibinfo{address}{Stockholmsm^^c3^^a4ssan,
  Stockholm Sweden}, \bibinfo{pages}{1939--1948}.
\newblock
\urldef\tempurl%
\url{http://proceedings.mlr.press/v80/hebert-johnson18a.html}
\showURL{%
\tempurl}


\bibitem[\protect\citeauthoryear{Kamiran and Calders}{Kamiran and
  Calders}{2012}]%
        {10.1007/s10115-011-0463-8}
\bibfield{author}{\bibinfo{person}{Faisal Kamiran} {and} \bibinfo{person}{Toon
  Calders}.} \bibinfo{year}{2012}\natexlab{}.
\newblock \showarticletitle{Data Preprocessing Techniques for Classification
  without Discrimination}.
\newblock \bibinfo{journal}{\emph{Knowl. Inf. Syst.}} \bibinfo{volume}{33},
  \bibinfo{number}{1} (\bibinfo{date}{Oct.} \bibinfo{year}{2012}),
  \bibinfo{pages}{1^^e2^^80^^9333}.
\newblock
\showISSN{0219-1377}
\urldef\tempurl%
\url{https://doi.org/10.1007/s10115-011-0463-8}
\showDOI{\tempurl}


\bibitem[\protect\citeauthoryear{Kamiran, Karim, and Zhang}{Kamiran
  et~al\mbox{.}}{2012}]%
        {6413831}
\bibfield{author}{\bibinfo{person}{Faisal Kamiran}, \bibinfo{person}{Asim
  Karim}, {and} \bibinfo{person}{Xiangliang Zhang}.}
  \bibinfo{year}{2012}\natexlab{}.
\newblock \showarticletitle{Decision Theory for Discrimination-Aware
  Classification}. In \bibinfo{booktitle}{\emph{Proceedings of the 2012 IEEE
  12th International Conference on Data Mining}} \emph{(\bibinfo{series}{ICDM
  '12})}. \bibinfo{publisher}{IEEE Computer Society}, \bibinfo{address}{USA},
  \bibinfo{pages}{924^^e2^^80^^93929}.
\newblock
\showISBNx{9780769549057}
\urldef\tempurl%
\url{https://doi.org/10.1109/ICDM.2012.45}
\showDOI{\tempurl}


\bibitem[\protect\citeauthoryear{Kamishima, Akaho, Asoh, and Sakuma}{Kamishima
  et~al\mbox{.}}{2012}]%
        {10.1007/978-3-642-33486-3_3}
\bibfield{author}{\bibinfo{person}{Toshihiro Kamishima},
  \bibinfo{person}{Shotaro Akaho}, \bibinfo{person}{Hideki Asoh}, {and}
  \bibinfo{person}{Jun Sakuma}.} \bibinfo{year}{2012}\natexlab{}.
\newblock \showarticletitle{Fairness-Aware Classifier with Prejudice Remover
  Regularizer}. In \bibinfo{booktitle}{\emph{Proceedings of the 2012th European
  Conference on Machine Learning and Knowledge Discovery in Databases - Volume
  Part II}} (Bristol, UK) \emph{(\bibinfo{series}{ECMLPKDD'12})}.
  \bibinfo{publisher}{Springer-Verlag}, \bibinfo{address}{Berlin, Heidelberg},
  \bibinfo{pages}{35^^e2^^80^^9350}.
\newblock
\showISBNx{9783642334856}


\bibitem[\protect\citeauthoryear{Kearns, Neel, Roth, and Wu}{Kearns
  et~al\mbox{.}}{2018}]%
        {pmlr-v80-kearns18a}
\bibfield{author}{\bibinfo{person}{Michael Kearns}, \bibinfo{person}{Seth
  Neel}, \bibinfo{person}{Aaron Roth}, {and} \bibinfo{person}{Zhiwei~Steven
  Wu}.} \bibinfo{year}{2018}\natexlab{}.
\newblock \showarticletitle{Preventing Fairness Gerrymandering: Auditing and
  Learning for Subgroup Fairness}. In \bibinfo{booktitle}{\emph{Proceedings of
  the 35th International Conference on Machine Learning}}
  \emph{(\bibinfo{series}{Proceedings of Machine Learning Research})},
  \bibfield{editor}{\bibinfo{person}{Jennifer Dy} {and}
  \bibinfo{person}{Andreas Krause}} (Eds.), Vol.~\bibinfo{volume}{80}.
  \bibinfo{publisher}{PMLR}, \bibinfo{address}{Stockholmsm^^c3^^a4ssan,
  Stockholm Sweden}, \bibinfo{pages}{2564--2572}.
\newblock
\urldef\tempurl%
\url{http://proceedings.mlr.press/v80/kearns18a.html}
\showURL{%
\tempurl}


\bibitem[\protect\citeauthoryear{Kearns, Neel, Roth, and Wu}{Kearns
  et~al\mbox{.}}{2019}]%
        {10.1145/3287560.3287592}
\bibfield{author}{\bibinfo{person}{Michael Kearns}, \bibinfo{person}{Seth
  Neel}, \bibinfo{person}{Aaron Roth}, {and} \bibinfo{person}{Zhiwei~Steven
  Wu}.} \bibinfo{year}{2019}\natexlab{}.
\newblock \showarticletitle{An Empirical Study of Rich Subgroup Fairness for
  Machine Learning}. In \bibinfo{booktitle}{\emph{Proceedings of the Conference
  on Fairness, Accountability, and Transparency}} (Atlanta, GA, USA)
  \emph{(\bibinfo{series}{FAT* ’19})}. \bibinfo{publisher}{Association for
  Computing Machinery}, \bibinfo{address}{New York, NY, USA},
  \bibinfo{pages}{100^^e2^^80^^93109}.
\newblock
\showISBNx{9781450361255}
\urldef\tempurl%
\url{https://doi.org/10.1145/3287560.3287592}
\showDOI{\tempurl}


\bibitem[\protect\citeauthoryear{Kim, Ghorbani, and Zou}{Kim
  et~al\mbox{.}}{2019}]%
        {Multiaccuracy10.1145/3306618.3314287}
\bibfield{author}{\bibinfo{person}{Michael~P. Kim}, \bibinfo{person}{Amirata
  Ghorbani}, {and} \bibinfo{person}{James Zou}.}
  \bibinfo{year}{2019}\natexlab{}.
\newblock \showarticletitle{Multiaccuracy: Black-Box Post-Processing for
  Fairness in Classification}. In \bibinfo{booktitle}{\emph{Proceedings of the
  2019 AAAI/ACM Conference on AI, Ethics, and Society}} (Honolulu, HI, USA)
  \emph{(\bibinfo{series}{AIES ’19})}. \bibinfo{publisher}{Association for
  Computing Machinery}, \bibinfo{address}{New York, NY, USA},
  \bibinfo{pages}{247^^e2^^80^^93254}.
\newblock
\showISBNx{9781450363242}
\urldef\tempurl%
\url{https://doi.org/10.1145/3306618.3314287}
\showDOI{\tempurl}


\bibitem[\protect\citeauthoryear{Propublica}{Propublica}{2020}]%
        {COMPASRe72:online}
\bibfield{author}{\bibinfo{person}{Propublica}.}
  \bibinfo{year}{2020}\natexlab{}.
\newblock \bibinfo{title}{COMPAS Recidivism Risk Score Data and Analysis}.
\newblock
\newblock
\newblock
\shownote{Retrieved September 18, 2020 from
  \url{https://www.propublica.org/datastore/dataset/compas-recidivism-risk-score-data-and-analysis}.}


\bibitem[\protect\citeauthoryear{Russell, Kusner, Loftus, and Silva}{Russell
  et~al\mbox{.}}{2017}]%
        {Russel-2017-WorldsCollide}
\bibfield{author}{\bibinfo{person}{Chris Russell}, \bibinfo{person}{Matt~J
  Kusner}, \bibinfo{person}{Joshua Loftus}, {and} \bibinfo{person}{Ricardo
  Silva}.} \bibinfo{year}{2017}\natexlab{}.
\newblock \showarticletitle{When Worlds Collide: Integrating Different
  Counterfactual Assumptions in Fairness}.
\newblock In \bibinfo{booktitle}{\emph{Advances in Neural Information
  Processing Systems 30}}, \bibfield{editor}{\bibinfo{person}{I.~Guyon},
  \bibinfo{person}{U.~V. Luxburg}, \bibinfo{person}{S.~Bengio},
  \bibinfo{person}{H.~Wallach}, \bibinfo{person}{R.~Fergus},
  \bibinfo{person}{S.~Vishwanathan}, {and} \bibinfo{person}{R.~Garnett}}
  (Eds.). \bibinfo{publisher}{Curran Associates, Inc.},
  \bibinfo{pages}{6414--6423}.
\newblock
\urldef\tempurl%
\url{http://papers.nips.cc/paper/7220-when-worlds-collide-integrating-different-counterfactual-assumptions-in-fairness.pdf}
\showURL{%
\tempurl}


\bibitem[\protect\citeauthoryear{Zafar, Valera, Gomez~Rodriguez, and
  Gummadi}{Zafar et~al\mbox{.}}{2017}]%
        {Zafar-2017-FairnessBeyondDTDI}
\bibfield{author}{\bibinfo{person}{Muhammad~Bilal Zafar},
  \bibinfo{person}{Isabel Valera}, \bibinfo{person}{Manuel Gomez~Rodriguez},
  {and} \bibinfo{person}{Krishna~P. Gummadi}.} \bibinfo{year}{2017}\natexlab{}.
\newblock \showarticletitle{Fairness Beyond Disparate Treatment \& Disparate
  Impact: Learning Classification without Disparate Mistreatment}. In
  \bibinfo{booktitle}{\emph{Proceedings of the 26th International Conference on
  World Wide Web}} (Perth, Australia) \emph{(\bibinfo{series}{WWW '17})}.
  \bibinfo{publisher}{International World Wide Web Conferences Steering
  Committee}, \bibinfo{address}{Republic and Canton of Geneva, CHE},
  \bibinfo{pages}{1171^^e2^^80^^931180}.
\newblock
\showISBNx{9781450349130}
\urldef\tempurl%
\url{https://doi.org/10.1145/3038912.3052660}
\showDOI{\tempurl}


\bibitem[\protect\citeauthoryear{Zhang, Lemoine, and Mitchell}{Zhang
  et~al\mbox{.}}{2018}]%
        {10.1145/3278721.3278779}
\bibfield{author}{\bibinfo{person}{Brian~Hu Zhang}, \bibinfo{person}{Blake
  Lemoine}, {and} \bibinfo{person}{Margaret Mitchell}.}
  \bibinfo{year}{2018}\natexlab{}.
\newblock \showarticletitle{Mitigating Unwanted Biases with Adversarial
  Learning}. In \bibinfo{booktitle}{\emph{Proceedings of the 2018 AAAI/ACM
  Conference on AI, Ethics, and Society}} (New Orleans, LA, USA)
  \emph{(\bibinfo{series}{AIES '18})}. \bibinfo{publisher}{Association for
  Computing Machinery}, \bibinfo{address}{New York, NY, USA},
  \bibinfo{pages}{335^^e2^^80^^93340}.
\newblock
\showISBNx{9781450360128}
\urldef\tempurl%
\url{https://doi.org/10.1145/3278721.3278779}
\showDOI{\tempurl}


\end{thebibliography}

\end{document}